\pdfoutput=1
\documentclass[twoside]{article}
\usepackage{graphicx}
\usepackage{subcaption}
\usepackage{amsmath,amssymb}
\usepackage{enumitem}
\usepackage{booktabs}   
\usepackage{multirow}   
\usepackage{caption}    
\usepackage{siunitx}    
\usepackage{bbm}
\usepackage{booktabs}
\usepackage{dsfont}
\usepackage{csvsimple}

\usepackage[accepted]{aistats2026}
\usepackage[round]{natbib}

\newcommand{\useSymbolFootnotes}{\renewcommand{\thefootnote}{\fnsymbol{footnote}}}
\newcommand{\useNumericFootnotes}{\renewcommand{\thefootnote}{\arabic{footnote}}}

\DeclareMathOperator{\softmax}{softmax}
\begin{document}

%
\runningtitle{Characterizing Distributional Shifts with Multi-Statistic Diffusion Trajectories}

%

\useSymbolFootnotes
\twocolumn[
\aistatstitle{Beyond Binary Out-of-Distribution Detection: Characterizing Distributional Shifts with Multi-Statistic Diffusion Trajectories}

\aistatsauthor{
  Achref Jaziri\footnotemark[1] \And
  Martin Rogmann \And
  Martin Mundt \And
  Visvanathan Ramesh
}

\aistatsaddress{
  Goethe University  \And
  Goethe University  \And
  University of Bremen \And
  Goethe University 
}
]
\footnotetext[1]{Correspondence: \texttt{Jaziri@em.uni-frankfurt.de }}

\useNumericFootnotes
\setcounter{footnote}{0}

\begin{abstract}
  Detecting out-of-distribution (OOD) data is critical for machine learning, be it for safety reasons or to enable open-ended learning. However, beyond mere detection, choosing an appropriate course of action typically hinges on the type of OOD data encountered. Unfortunately, the latter is generally not distinguished in practice, as modern OOD detection methods collapse distributional shifts into single scalar outlier scores. This work argues that scalar-based methods are thus insufficient for OOD data to be properly contextualized and prospectively exploited, a limitation we overcome with the introduction of DISC: Diffusion-based Statistical Characterization. DISC leverages the iterative denoising process of diffusion models to extract a rich, multi-dimensional feature vector that captures statistical discrepancies across multiple noise levels. Extensive experiments on image and tabular benchmarks show that DISC matches or surpasses state-of-the-art detectors for OOD detection and, crucially, also classifies OOD type, a capability largely absent from prior work. As such, our work enables a shift from simple binary OOD detection to a more granular detection.
\end{abstract}

\section{INTRODUCTION}

In the real world, distribution shifts are the norm rather than the exception \citep{quinonero2022dataset}. Predictive models are commonly used on test data which are drawn from distributions that differ from the training data, such as images acquired with a different protocol or a population change \citep{amodei2016concrete}. These settings highlight the importance of building models that are robust to natural transformations \citep{hendrycks2018deep}. Two prevalent strategies exist for dealing with distribution shifts: improving model generalization or detecting the shifts themselves \citep{zhou2022domain, hendrycks2016baseline}. In this context, detecting out-of-distribution (OOD) data has become increasingly popular by training supervised predictive models or generative models on in-distribution data and examining some proxy features. 

However, not all flagged shifts are and should be treated equally. Some shifts may indicate corrupted or irrelevant data that should be discarded, while others may represent opportunities for model adaptation and continual learning \citep{Mundt2023wholistic,wang2024comprehensive}. This raises a critical question: \emph{are current OOD detection methods able to distinguish between different types of distribution shifts beyond the mere detection of a shift's existence?}

For intuition, consider a standard image setting with CIFAR-10 \citep{cifar10} as in-distribution and two OOD regimes: covariate corruptions and semantic shifts (e.g., unseen classes like butterflies). While several methods perform very well in the standard ID vs OOD setting, they remain fundamentally limited in their ability to distinguish between different types of shifts, a point we will quantify later. This is in part  because scalar OOD scores conflate heterogeneous shifts. For instance, the theoretically sound family of open-set recognition methods aims to distinguish OOD samples by quantifying bounds on the "known" space, i.e., the closed set \citep{boult2019learning}. These approaches are typically brought into practice by defining compact abating probabilities—measures that decay probabilities of decisions towards zero when moving beyond the training data support \citep{scheirer2012toward, scheirer2014probability}. In similar spirit, generative approaches predominantly employ measures of statistical (dis-)similarity, decaying the likelihood of data 
away from the fitted training density \citep{kirsch2021test}. By definition, either of these approaches considers all OOD samples to be fundamentally the same, failing to capture the nuanced differences between various types of distributional shifts (see Figure \ref{intro_fig} as an example).

\begin{center}
\includegraphics[width=\linewidth]{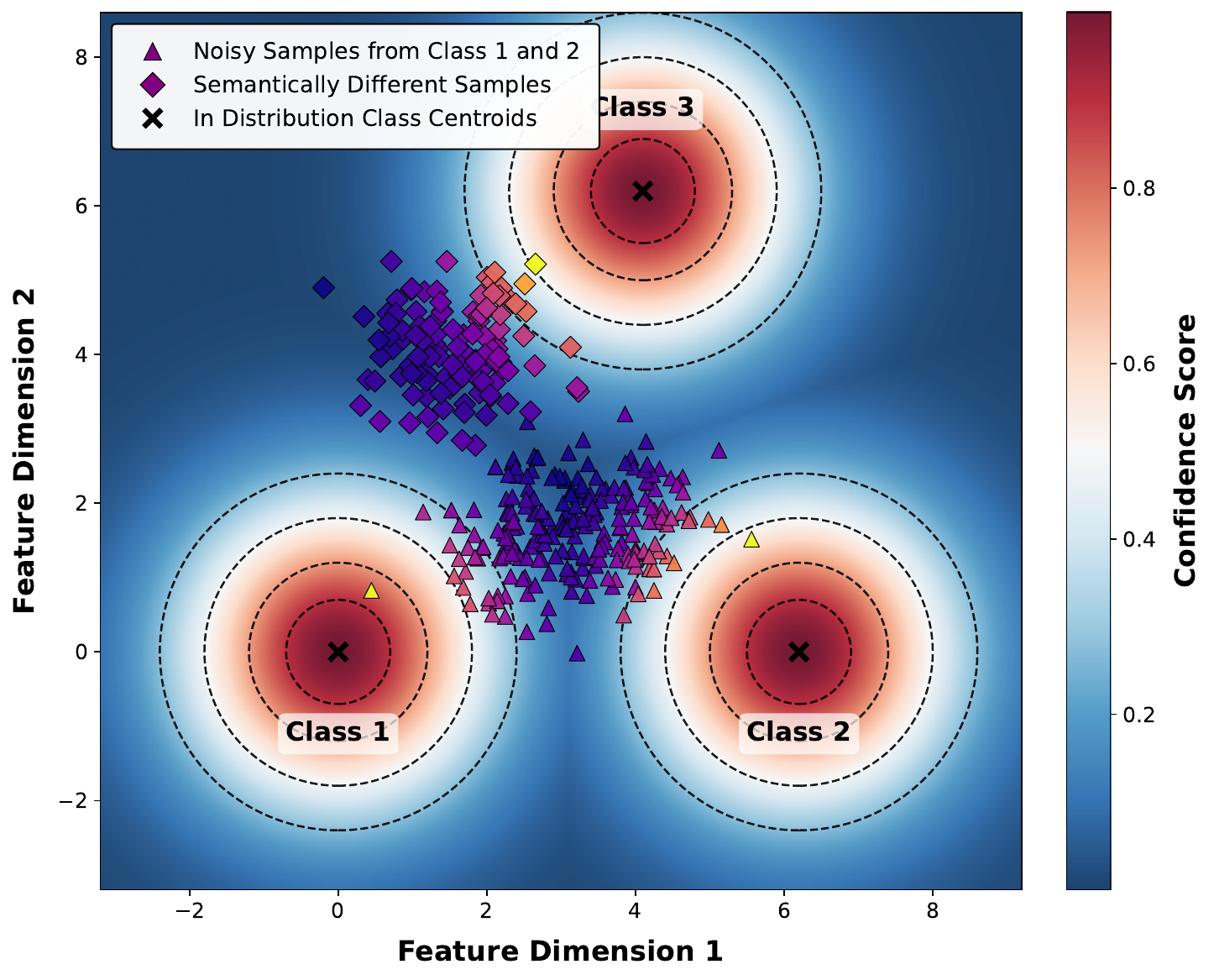}
\captionof{figure}{Simplified illustration to highlight that standard out-of-distribution (OOD) detection methods typically fail to differentiate between the distinct types of shift. In the example, density or distance-based methods would collapse their scores and conflate the two kinds of OOD samples. Reasoning of failure modes for other methods follows analogously.} 
\label{intro_fig}
\end{center}

This work thus seeks to move beyond the traditional binary classification of inputs as simply "in" or "out" of distribution. 
We first corroborate above mentioned observations empirically and show that the difficulty in distinguishing between different types of OOD stems from common limitations of existing methods. We further demonstrate theoretically that this limitation is inherent to any approach using a single test statistic, as it necessarily collapses the multidimensional nature of distributional differences into a single dimension. 
Finally, we propose a method based on denoising diffusion probabilistic models (DDPMs) \citep{ho2020denoising} that addresses these limitations. Our introduced scoring approach captures discrepancies at multiple levels of abstraction, enabling us to disentangle different types of distributional shifts for different data modalities.

\section{PRELIMINARIES AND BACKGROUND}
We formalize the taxonomy used throughout the paper. The following subsections fix notation, define shift types, and review OOD detection approaches. 

\subsection{Preliminaries: OOD Detection}
\looseness=-1
We assume a standard supervised learning setting in which each training pair \(\bigl(x_{i},y_{i}\bigr)\) is drawn i.i.d.\ from a joint distribution  $  P_{\mathrm{id}}(X,Y)\quad\text{over}\quad
  \mathcal{X}_{\mathrm{id}}\times \mathcal{Y}_{\mathrm{id}}$,
where  \(\mathcal{X}_{\mathrm{id}}\subset \mathcal{X}\) is the input space of in-distribution (ID) data and \(\mathcal{Y}_{\mathrm{id}}=\{1,2,\dots,K\}\) is the set of known class labels. We denote by \(P=P_{\mathrm{id}}\) the marginal distribution of \(X\) restricted to \(\mathcal{X}_{\mathrm{id}}\). In its most general form, OOD detection can be framed as a single-sample hypothesis test \citep{nalisnick2018deep}.  Let \(\mathcal{Q}\) denote a family of candidate out-of-distribution alternatives, with \(P \notin \mathcal{Q}\). Given a sample \(x\), the goal is to decide between \(H_0: x \sim P\) and \(H_A: x \sim Q,\; Q \in \mathcal{Q}\).

The decision is based on a predetermined test statistic  $\phi:\mathcal{X}\;\to\;\mathbb{R}$
which can be any (measurable) function of the single sample \(x\).  Commonly, one rejects \(H_{0}\) (i.e.\ declares \(x\) to be OOD) if 
$\phi(x) \;<\; k$ for some threshold \(k\).   In practice, one may define a test statistic $\phi(x)$ and reject \(H_{0}\) whenever \(\phi(x)<\tau\).  The goal is that \(\phi(x)\) be well‐separated on \(\mathcal{X}_{\mathrm{id}}\) versus \(\mathcal{X}_{\mathrm{ood}}\), so that novel or shifted inputs are flagged as OOD with high reliability.

\subsection{Types of Out-of-Distribution Shifts}

The space of out-of-distribution inputs $\mathcal{X}_{\text{ood}} = \mathcal{X} \setminus \mathcal{X}_{\text{id}}$ can be decomposed into several subsets that correspond to distinct types of distributional shifts. These shifts affect either the marginal distribution of inputs, the label distribution, or the semantic content as prominently defined by Quinonero et. al  \citep{quinonero2022dataset}. 

In this work, we focus on covariate and semantic shifts because they act directly on the input distribution, either altering its geometry/texture or introducing novel concepts. We thus most directly test an OOD detector’s ability to reason about features and semantics at inference time.  Label shifts primarily reflect downstream curation effects, such as class reweighting, later editing, or mislabeling, rather than changes in the support or geometry of the input distribution. They also require different assumptions and estimation tools and are therefore beyond our present scope.

\textbf{Covariate Shift} occurs when the distribution of inputs changes while the conditional label distribution remains unchanged. In this case, the support of $\mathcal{X}_{\text{test}}$ includes regions that are disjoint from $\mathcal{X}_{\text{id}}$, i.e., $\mathcal{X}_{\text{ood}}^{\text{cov}} \subset \mathcal{X}_{\text{ood}}$. Examples include shifts in background, sensor characteristics, or adversarial perturbations.

\textbf{Semantic Shift} denotes the appearance of novel classes not seen during training. In this case, test inputs from these unseen classes lie in a subset $\mathcal{X}_{\text{ood}}^{\text{sem}} \subset \mathcal{X}_{\text{ood}}$, and their associated labels belong to $\mathcal{Y}_{\text{ood}} = \mathcal{Y}_{\text{test}} \setminus \mathcal{Y}_{\text{id}}$. This is the canonical setting for open-set recognition and novelty detection \citep{scheirer2012toward, pimentel2014review}.

Our primary emphasis is on \emph{differences within} a given shift type, rather than only contrasting covariate versus semantic shifts as a whole, as interventions and risks differ across families even within the same type (e.g., in continual settings: denoise vs. throw away vs. retrain). We therefore treat each corruption or dataset as a distinct OOD family.  

\begin{figure*}[t!]
  \centering
  \begin{subfigure}[t]{0.32\textwidth}
    \centering
    \includegraphics[width=\linewidth]{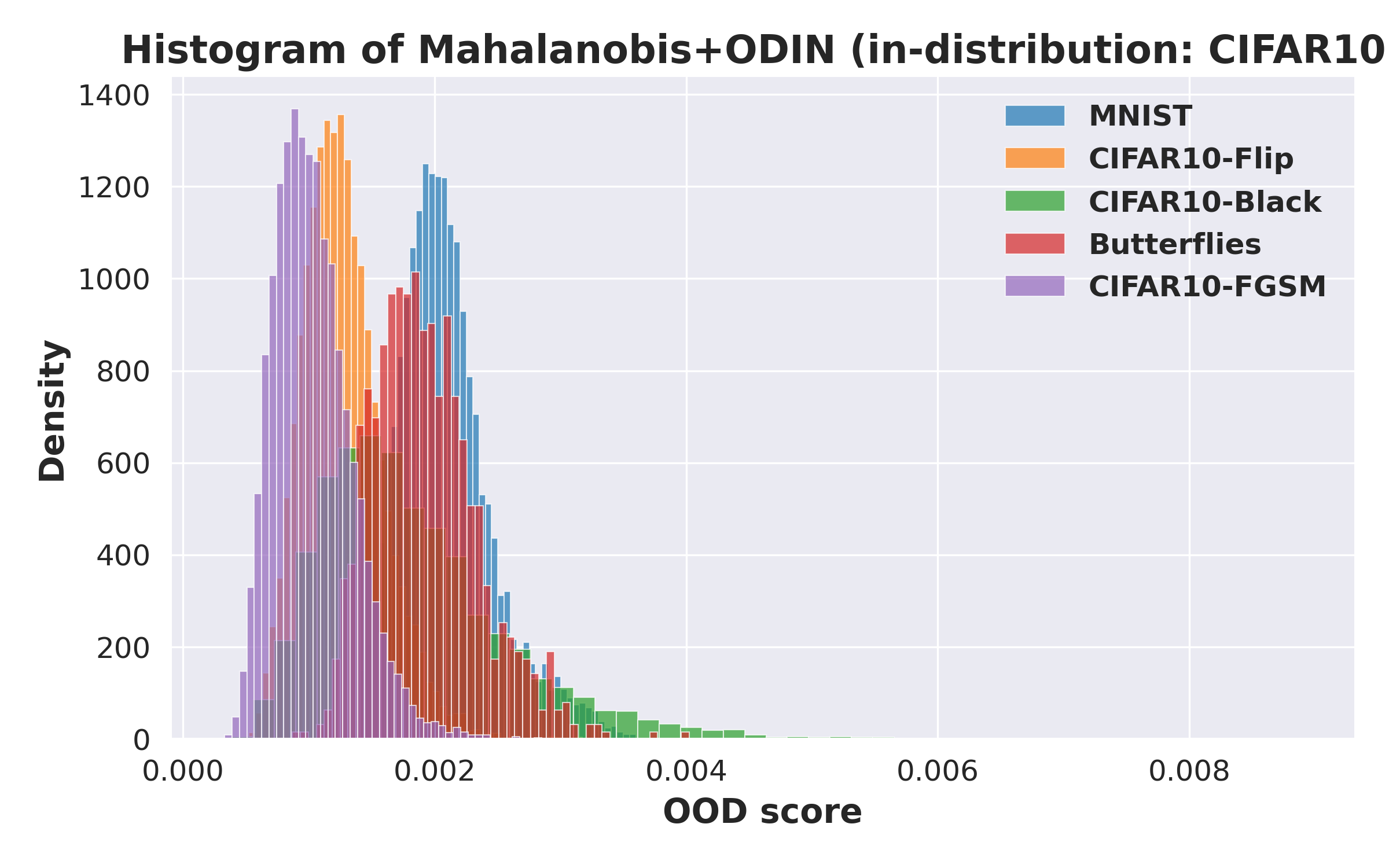}
    \caption{Mahalanobis+ODIN}
    \label{fig:imageNet:mahalanobis}
  \end{subfigure}
  \hfill
  \begin{subfigure}[t]{0.32\textwidth}
    \centering
    \includegraphics[width=\linewidth]{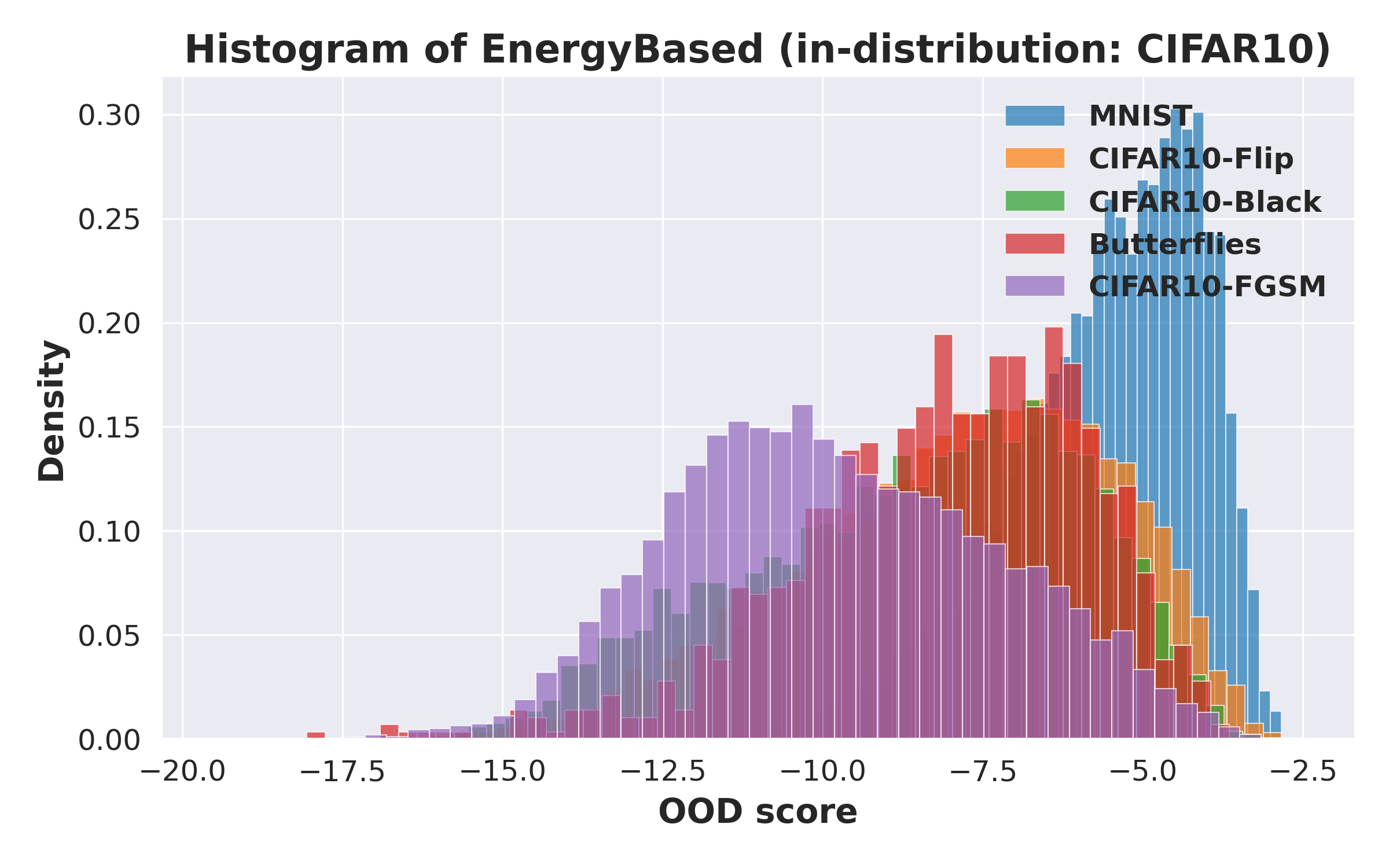}
    \caption{Energy-based}
    \label{fig:imageNet:odin}
  \end{subfigure}
  \hfill
  \begin{subfigure}[t]{0.32\textwidth}
    \centering
    \includegraphics[width=\linewidth]{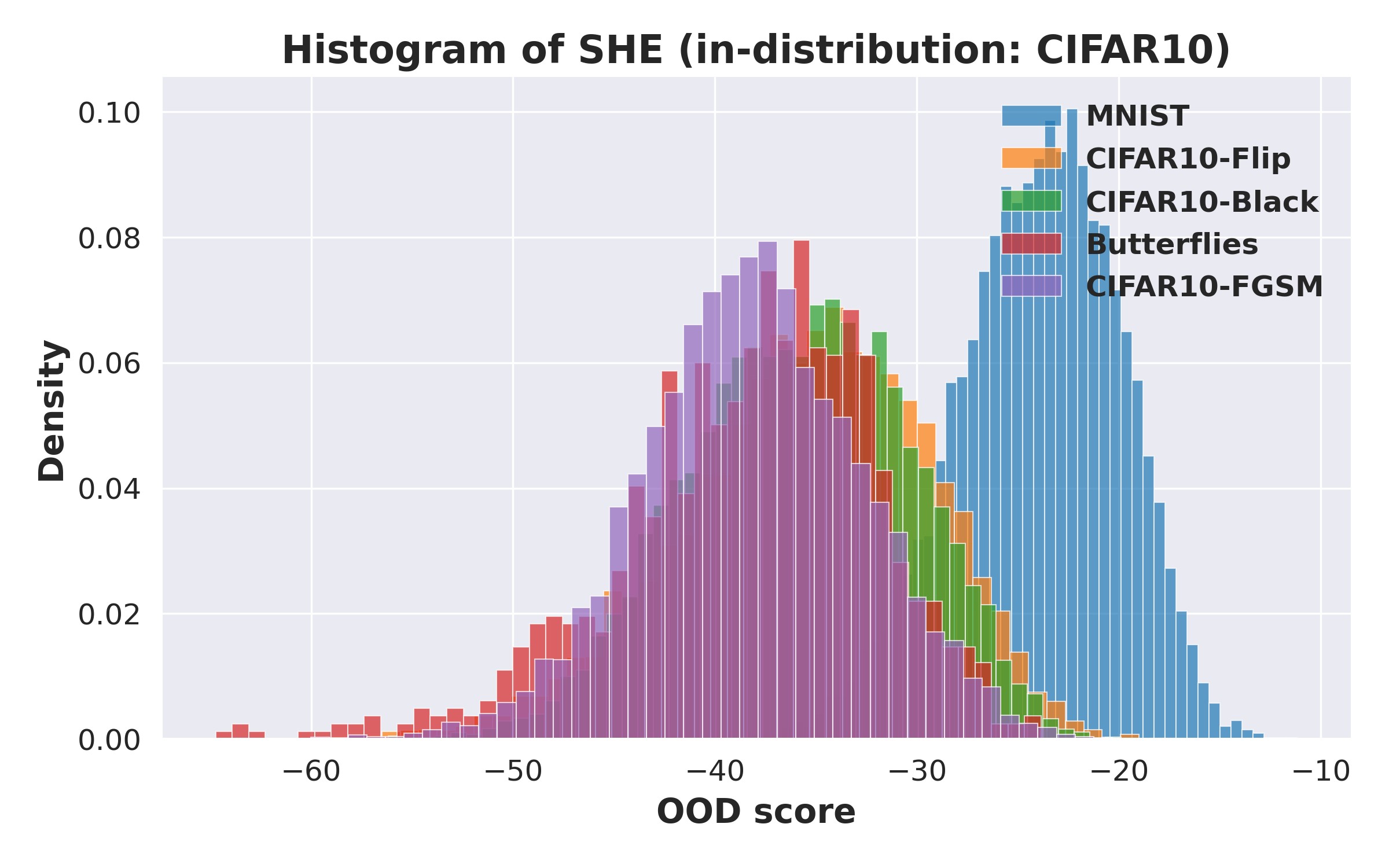}
    \caption{SHE}
    \label{fig:cifar:she}
  \end{subfigure}
  
  \caption{OOD scores overlap significantly for different OOD types for models trained on CIFAR-10 as in-distribution. This observation is independent of the employed OOD detector (panels a-c).}
  \label{fig:CIFAR:score_distributions}
\end{figure*}

\subsection{OOD Score Computation}\label{app:ood-taxonomy}
Regardless of the modeling choice, most modern OOD detectors compress the evidence for an input $x$ into a \emph{single} scalar and apply a common decision rule,
$
\delta(x)=\mathds{1}\!\left[o(x)\ge \tau\right].
$
What differs is \emph{where} this evidence comes from: the prediction space, the representation space, or a generative model, and whether it is further reshaped at test time. We categorize representative methods through this lens:
\textbf{Output–score based.} These methods derive $o(x)$ from classifier outputs without modeling density. Canonical choices include maximum softmax probability (MSP), predictive entropy \citep{hendrycks2016baseline}, and the energy score $o_E(x)=-\log\sum_c \exp f_c(x)$. The latter avoids softmax normalization and often improves separation \citep{liu2020energy}. Calibration can be improved by ODIN (temperature scaling + small input perturbations) or Outlier Exposure (OE), which penalizes overconfidence on auxiliary outliers \citep{hsu2020generalized,hendrycks2018deep}. Ensembles and MC-Dropout further encourage uncertainty on unfamiliar inputs \citep{lakshminarayanan2017simple,gal2016dropout}.

\textbf{Feature–based.} These methods compute $o(x)$ from layer embeddings $z(x)$. Class-conditional (Mahalanobis) distances and related prototype distances measure how far $z(x)$ lies from ID clusters; nearest-neighbor variants estimate isolation in feature space \citep{lee2018simple,ren2021simple,sun2022out}. Activation-shaping improves these signals at test time: ReAct truncates extreme activations, DICE clips activations dynamically, and ASH uses asymmetric clipping; each can be paired with logit- or feature-based scores \citep{sun2021react,sun2022dice,djurisic2022extremely}. Gradient/sensitivity cues (e.g., GradNorm) and Fisher-style approximations also correlate with being off-manifold \citep{huang2021importance,dauncey2024approximations}.

\textbf{Generative: density and reconstruction methods.}
Explicit likelihood models (PixelCNN++, Glow) set $o(x)=-\log p(x)$ \citep{salimans2017pixelcnn,kingma2018glow}. However, raw likelihood can mis-rank OOD due to sensitivity to low-level statistics \citep{nalisnick2018deep,zhang2021understanding}. Reconstruction methods (VAEs; diffusion) score via $\lVert x-r(x)\rVert$, or via typicality surrogates. Recent diffusion-based detectors leverage denoising trajectories, inpainting errors, and noisy–denoised feature mismatches to boost performance without labels \citep{graham2023denoising,livernoche2023diffusion,liu2023unsupervised,heng2024out,yang2024diffusion,oko2023diffusion}.

\textbf{Inference-modifying and augmentation-based.}
Beyond fixed scoring, many methods perturb inputs or internal activations to widen ID/OOD separation at test time—ODIN-style input nudges or activation clipping are typical examples \citep{hsu2020generalized,sun2021react,sun2022dice}.

Strong systems often \emph{combine} components (e.g., activation shaping + energy; Mahalanobis on features from a diffusion backbone). Yet there are trade-offs between “near” and “far” OOD, and no single score performs well across all regimes \citep{guille2024expecting,zhang2021understanding,li2025out}. Finite-sample effects and high dimensionality further complicate separation \citep{ghosal2024overcome,aggarwal2001surprising}. Recent impossibility results show OOD detection is not distribution-free PAC-learnable in general; learnability requires structure relating ID and OOD or restrictions on hypothesis spaces \citep{fang2022out,garov2025closer}.

\section{MOTIVATING OBSERVATIONS}
In this section, we first present the empirical observations that motivate our work, then link them to design choices in current OOD detectors, and finally corroborate them with a theoretical non-identifiability result showing the limits of standard OOD tests.

\subsection{Score Distributions Align Across OODs}

\begin{figure}[t!]
    \includegraphics[width=\linewidth]{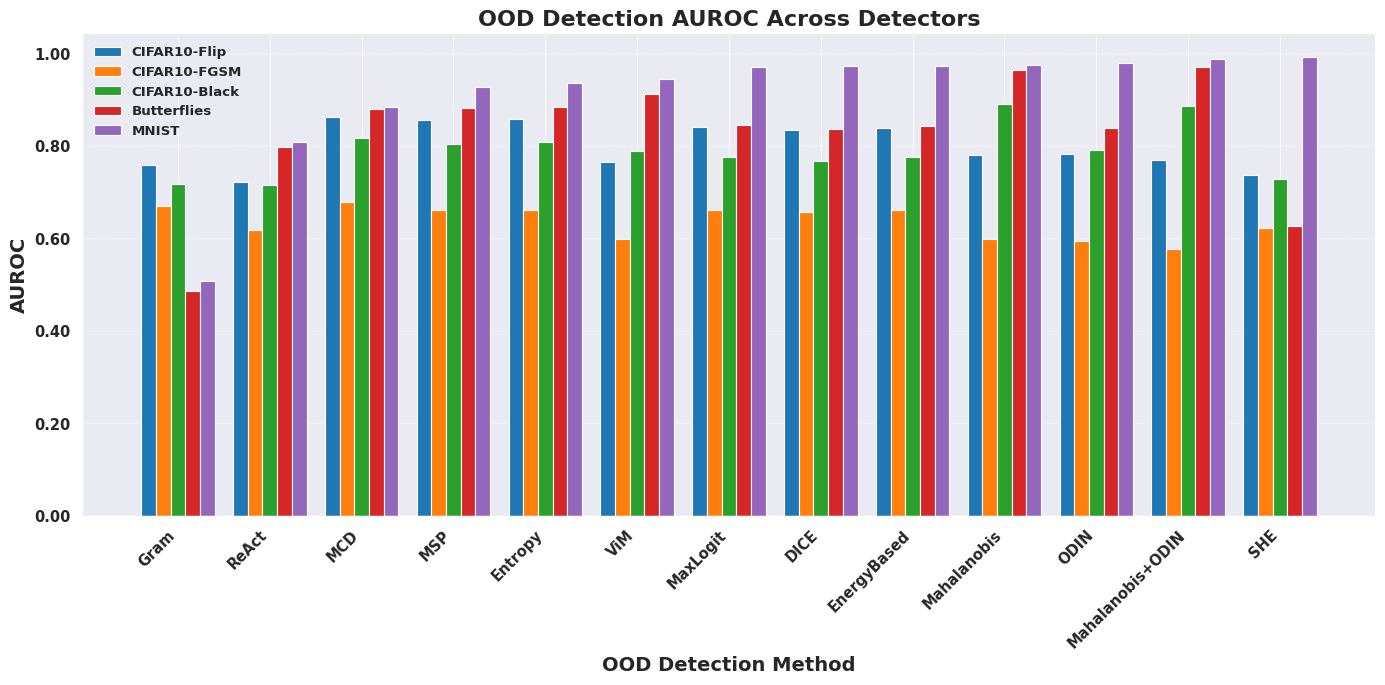}
  \caption{ 
 AUROC in the standard OOD detection setting across various OOD types remains high across detectors (x-axis) for models trained on CIFAR-10.
}
  \label{fig:cifar:auroc}
\end{figure}

\begin{figure}
    \centering
    \includegraphics[width=\linewidth]{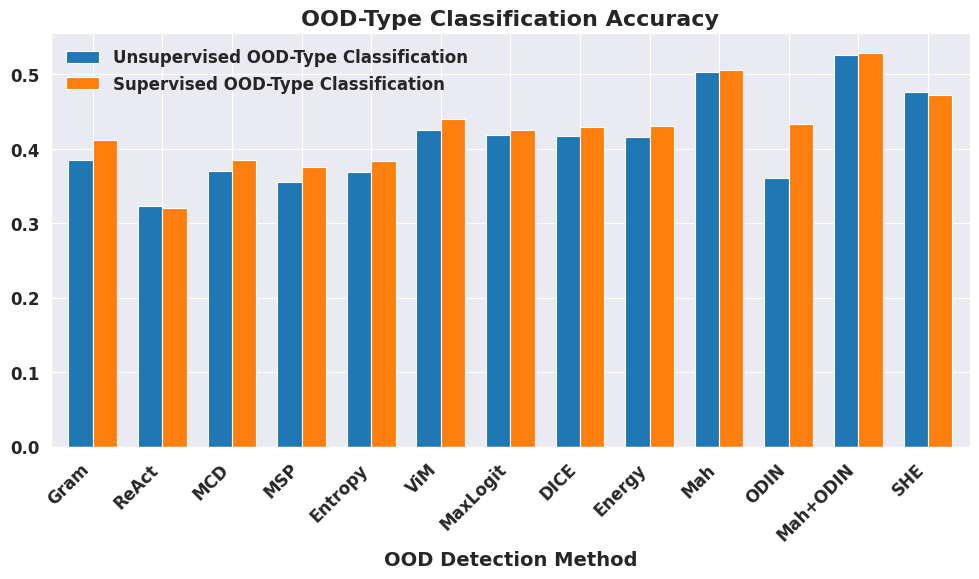}
  \caption{OOD-type classification accuracy (using unsupervised K-means and supervised MLP approaches) is substantially worse when compared to traditional binary OOD-detection, as presented in Figure \ref{fig:cifar:auroc}.}
  \label{fig:cifar:ood_types}
\end{figure}

To examine how existing OOD detectors respond to diverse out-of-distribution inputs, we analyze OOD detection scores for multiple datasets on a shared CIFAR-10 in-distribution base. Figure~\ref{fig:CIFAR:score_distributions} displays representative histograms for Mahalanobis (with and without ODIN), Energy-based scoring, and SHE \citep{zhang2022out}. Across methods, each OOD family produces a clear shift away from the ID mode; however, the OOD histograms \emph{overlap heavily} with one another. Consistent with this, standard ID--vs--OOD AUROC remains high across detectors (Figure ~\ref{fig:cifar:auroc}), When we ask \emph{which} OOD family a sample belongs to, accuracy drops markedly in both unsupervised (K-means) and supervised (MLP) settings (Figure~\ref{fig:cifar:ood_types}).

The \emph{relative} positions of the different OOD types are also strikingly similar across methods: \texttt{CIFAR-FGSM} and \texttt{CIFAR-Flip} cluster closer to the ID peak, whereas \texttt{Butterflies} and \texttt{MNIST} are scored as more novel. This consistency holds for logit-based (Energy), feature-distance (Mahalanobis), and score-normalized embedding (SHE) approaches.



\subsection{Constraints of Current OOD Detectors}

The empirical regularity found in the previous subsection points to a deeper issue: most existing approaches rely on scoring functions that are inherently scalar and coarse. They fail to disentangle between different shifts. Fundamentally, these scalar approaches collapse the multidimensional nature of distributional differences into a single dimension, typically following some form of exponential or heavy-tailed distribution that tends to be unimodal. This dimensionality reduction pushes all different types of OOD inputs toward the extremes or tails of the score distribution, making distinguishing between them impossible regardless of their underlying structural differences.

Moreover, supervised models are trained to minimize classification error on in-distribution data and are not optimized to assess distributional membership. As a result, they conflate discriminative uncertainty (i.e., uncertainty over class labels) with epistemic uncertainty about the underlying data distribution. This conflation induces irreducible errors, particularly in near-OOD regimes, where samples lie in ambiguous feature regions or closely resemble in-distribution examples. However, the distance to the boundary does not tell us the nature of the shift.

Generative models aim to capture the data distribution $p(x)$ and are thus often viewed as more principled for OOD detection. However, methods that rely solely on the likelihoods produced by these models inherit their own fundamental limitations. The quantity $p(x)$ does not provide direct access to the decision-relevant posteriors $p(\text{OOD}_i \mid x)$, where $\text{OOD}_i$ denotes a specific type of out-distribution. In other words, a high likelihood under $p(x)$ does not imply low novelty across all possible modes of OOD variation, nor does it allow discrimination between different types of OOD inputs.


Typicality-based methods attempt to mitigate this mismatch by testing whether a sample lies in a high-probability region of the estimated distribution $p(x)$, rather than relying solely on its likelihood value. However, these methods still reduce the input to a scalar test statistic $\phi_p(x): \mathcal{X} \to \mathbb{R}$, using some sort of exponential distribution, effectively projecting the high-dimensional input space onto a one-dimensional axis. 

\subsection{Non-Identifiability Under Singular Test-stastic: A Theoretical Result} %

Prior work \citep{zhang2021understanding} formalizes a fundamental limitation of scalar-valued test statistics in the context of single-sample out-of-distribution detection. They show that for any test statistic \( \phi_p: \mathcal{X} \to \mathbb{R} \), if the conditional distribution \( P(X \mid \phi_p(X) = t) \) is non-degenerate on a set of nonzero measure, then there exists an alternative distribution \( Q \neq P \) such that \( \phi_p(X) \sim Q = \phi_p(X) \sim P \). Consequently, any test relying solely on \( \phi_p(x) \) cannot distinguish between \( P \) and such alternatives, and has power equal to the false positive rate. Extending this proposition, we show that multiple, distinct distributions can induce the same distribution over \( \phi_p(X) \), rendering them mutually indistinguishable under any test based solely on \( \phi_p(x) \):

\paragraph{Proposition.} \emph{Let \(P\) be a probability measure on a sample space \(\mathcal X\) and let 
\(\phi_{p}\colon\mathcal X\to\mathbb R\) be any (measurable) test statistic.  
Suppose there exists a measurable set  $\Phi_{0}\subseteq\mathrm{Range}(\phi_{p})$ with $\quad
  P\!\bigl(\phi_{p}(X)\in\Phi_{0}\bigr)\;>\;0$
such that, for almost every \(\phi\in\Phi_{0}\), the conditional law 
\(P(\,\cdot\mid\phi_{p}(X)=\phi)\) is \emph{non-degenerate} (i.e.\ not a point mass). Then, there exist two distributions \( Q_1 \) and \( Q_2 \) such that \( Q_1 \neq P \), \( Q_2 \neq P \), and \( Q_1 \neq Q_2 \), yet all three distributions induce the same distribution over \( \phi_p(X) \); specifically,}
$
\phi_p(X) \sim Q_1 = \phi_p(X) \sim Q_2.
$
\emph{As a result, any test that relies solely on the value of \( \phi_p(x) \) has power equal to the false positive rate, and therefore cannot distinguish between \( Q_1 \) and \( Q_2 \).}

\paragraph{Proof Sketch.} See the Appendix A for the full proof. The proof sketch is as follows: We construct two distributions, \(Q_1\) and \(Q_2\), that are distinct from \(P\) and each other, yet produce the exact same marginal distribution for the statistic \(\phi_p(X)\). The strategy is to identify a set of statistic values, \(\Phi\), where the conditional distribution \(P(X \mid \phi_p(X) = \phi)\) is not concentrated on a single point (non-degenerate Condition). For each \(\phi \in \Phi\), this non-degeneracy allows us to find at least two disjoint subsets of the outcome space, \(A_\phi\) and \(B_\phi\), which both have positive conditional probability. We then define new conditional distributions, \(q_1(x \mid \phi)\) and \(q_2(x \mid \phi)\). For \(\phi \in \Phi\), these new distributions slightly re-weight the probability mass between the sets \(A_\phi\) and \(B_\phi\). For any \(\phi \notin \Phi\), we leave the conditional distribution unchanged \(\bigl(q_i(x \mid \phi) = p(x \mid \phi)\bigr)\). Finally, we construct the full distributions \(Q_1\) and \(Q_2\) by combining these new conditional distributions with the original marginal density of \(\phi_p(X)\). This procedure guarantees that the marginal law of \(\phi_p(X)\) is identical under \(P\), \(Q_1\), and \(Q_2\), while the joint distributions are provably different. Any test that depends only on $\phi_{p}(X)$ therefore attains identical
rejection probabilities under all three measures, so its power equals its
false-positive rate.

\section{SCORE-BASED GENERATIVE MODELS FOR MULTI-DIMENSIONAL OOD DETECTION}

In this section, we first motivate the use of score-based generative models to expose structure across noise levels. We then introduce our suggested approach, \emph{DISC}, which replaces a single anomaly score with a multi-statistic embedding. We conclude by specifying the metric suite and feature construction for both images and tabular data.

\subsection{Motivation}

A fundamental challenge in OOD detection is that the separability of in- and out-of-distribution samples depends on where in the pipeline one looks—input, latent, or output space. Certain classes of OOD samples can appear indistinguishable in one space but clearly deviate in another. For example, adversarial samples are nearly identical to normal ones in input space but induce drastically different activations in latent and output space. Conversely, perturbations like adding a few black pixels produce negligible changes in latent or output representations, despite being clearly anomalous at the input level. More generally, if a model is invariant to the features responsible for a given shift, then examining a single representational space will inevitably collapse certain classes of OOD. This observation explains why adversarial detection methods have often been treated as a separate research branch \citep{metzen2017detecting}. 

\begin{figure}[t]
    \centering
    \begin{subfigure}[t]{0.48\textwidth}
        \centering
        \includegraphics[width=\linewidth]{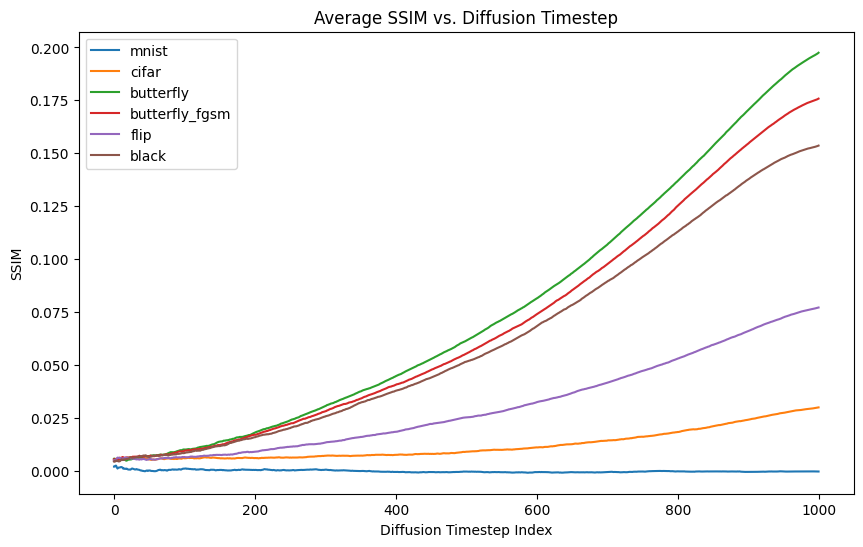}
        \caption{}
        \label{fig:ssim}
    \end{subfigure}
    \hfill
    \begin{subfigure}[t]{0.48\textwidth}
        \centering
        \includegraphics[width=\linewidth]{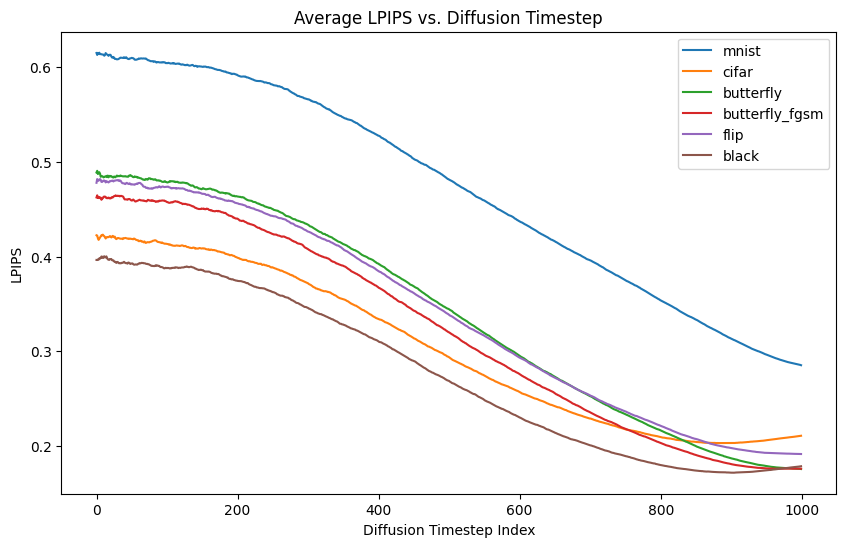}
        \caption{}
        \label{fig:mse}
    \end{subfigure}

    \vspace{0.5cm}
    \caption{(a) Average SSIM vs. diffusion timestep. When trained on butterflies, SSIM highlights the structural separation of naturalistic shifts (e.g., flip, black occlusion) from in-distribution samples as denoising progresses. (b) Average LPIPS vs. diffusion timestep. Perceptual similarity score for different distribution shifts show different trends over time, revealing the importance of time-expanded information in capturing differences across OOD types.}
    \label{ButtefliesExample}
\end{figure}

The limitations of scalar-based detection methods motivate a shift toward richer, multi-dimensional characterizations of distributional shift.  We adopt score-based generative models \citep{song2020score} as the foundation for this approach: \emph{their iterative denoising process decomposes data generation across multiple timesteps, naturally exposing both coarse semantic structure and fine-grained statistical details}. \\
To motivate with an example (Figure \ref{ButtefliesExample}): when training on butterflies, the Structural Similarity Index (SSIM) \citep{wang2004image} separates naturalistic shifts such as flip or black occlusion from in-distribution samples, reflecting divergence in structural coherence. A perceptual metric like LPIPS \citep{zhang2018unreasonable}, on the other hand, highlights a complementary perspective: perceptual similarity decreases more sharply for certain datasets (e.g., CIFAR) while remaining comparatively stable for others (e.g., black occlusion), with these contrasts becoming more pronounced at later timesteps. Taken together, SSIM and LPIPS illustrate a complementary effect that motivates the design of DISC, which explicitly integrates diverse metrics across the diffusion trajectory to achieve a more comprehensive characterization of distributional shift. 

\subsection{DISC: DIffusion‐based Statistical Characterization for OOD Detection}

Score-based generative models \citep{sohl2015deep, song2020score} learn to approximate the score function \(\nabla \log p(x)\) by training a family of denoisers \(\{D_{\sigma}\}\) at varying noise levels.  Beyond generative sampling, score-based generative models offer a principled framework for constructing a collection of test statistics that address the limitations of scalar-valued detectors.

Given a clean sample \(x_0 \sim p(x)\), Gaussian perturbations produce noisy versions \(x_{\sigma} \sim \mathcal{N}(x_0, \sigma^2 I)\).  At each noise level \( \sigma \), the model learns a denoiser \( D_{\sigma}(x_{\sigma}) \) that estimates the posterior mean \( \mathbb{E}[x \mid x_{\sigma}] \), or equivalently, the score function \( \nabla \log p_{\sigma}(x_{\sigma}) \) via Tweedie’s formula  \citep{efron2011tweedie}. This leads to a family of estimators \( \{D_{\sigma}\}_{\sigma \in \Sigma} \), each associated with a different degree of corruption and thus encoding distinct information.

At sampling time, these learned score functions are employed to reverse the diffusion process and generate samples from the underlying clean distribution \(p(x)\) via a discretized reverse diffusion procedure.

This decomposition across noise levels naturally induces a set of test statistics, each conditioned on \( \sigma \), which compare forward-diffused samples \( x_{\sigma} \sim p(x_{\sigma} \mid x) \) to their corresponding reconstructions \( D_{\sigma}(x_{\sigma}) \).

Rather than relying on a single test statistic \(\phi(x)\), we propose a more general \emph{multi‐statistic framework} that leverages a collection of complementary metrics.  For each noise level \(\sigma\in\Sigma\), we define
\[
\phi_\sigma^{(m)}(x)\;=\;d^{(m)}\bigl(x_\sigma,\;D_\sigma(x_\sigma)\bigr),
\quad m=1,\dots,M,
\]
where each discrepancy \(d^{(m)}\) captures a distinct view of the data manifold (e.g.\ pixel fidelity, perceptual quality, structural integrity, or distributional statistics).  
Aggregating these \(M\) metrics across all \(N=|\Sigma|\) levels yields the embedding
\[
s(x)
=
\bigl[
\{\phi_{\sigma_i}^{(m)}(x)\}_{m=1}^{M}
\bigr]_{i=1}^{N}
\in \mathbb{R}^{NM}.
\]
which serves as the input for various downstream OOD tasks. As mentioned, we refer to this framework as DISC: Diffusion-based Statistical Characterization.

\subsection{Complementary Metrics to Distinguish Different Types of OOD}
The central question is how to select a collection of test statistics that enables \emph{fine-grained} OOD detection. The choice of metrics depends on the data modality and on the specific shifts we aim to disentangle. Our guiding principle is to probe complementary manifestations of shift—pixel-level deviations, local structure,  perceptual change, representation stability, and texture/frequency statistics. The goal is a \emph{partially orthogonal} set rather than a single “best” score.

At each noise level \(\sigma_i\), we evaluate four metrics on the noisy input \(x_{\sigma_i}\) and its denoised estimate \(\hat{x}_{\sigma_i}\): Mean Squared Error (MSE) captures pixel-level deviations; Learned Perceptual Image Patch Similarity (LPIPS) \citep{zhang2018unreasonable} measures discrepancies in deep feature space, reflecting higher-level, perceptual changes; the Structural Similarity Index (SSIM) \citep{wang2004image} assesses local luminance, contrast, and texture fidelity; and Local Complexity (LC) \citep{humayun2024deep, humayun2024secrets} quantifies the stability of intermediate representations under small perturbations.

Metrics computed in learned feature spaces (e.g., perceptual distances) inherit the invariances of their networks and training datasets. While beneficial for the in-distribution, these invariances can create "blind spots" \citep{li2025out}: if the model down-weights texture, fine-scale frequency content, or mild geometric warps, distances in that space may not detect shifts expressed along those factors. To mitigate this, we pair learned metrics with specification-driven, \emph{non-learned} descriptors (in this case texture and frequency statistics). The resulting trajectory vector is more disentangled, less sensitive to dataset-specific invariances.

Texture and frequency descriptors (e.g., Local Binary Patterns \citep{pietikainen2010local} and Discrete Wavelet Transform bands \citep{heil1989continuous}) provide orthogonal information by capturing fine-grained patterns, spectral energy distributions, and orientation structure that are often invisible to pixel- or feature-level distances \citep{jaziri2024representation, ditzel2025uncertainty}. Accordingly, in parallel we extract these descriptors from \(x_{\sigma_i}\) and \(\hat{x}_{\sigma_i}\). For each descriptor type, we construct normalized histograms \(h_{\sigma_i}\) and \(\hat{h}_{\sigma_i}\) with \(K\) bins per sample and \(\epsilon\)-smoothing for numerical stability, and compare them using the Kullback–Leibler divergence. These divergences are included in $s(x)$.

For tabular data, where perceptual and texture descriptors are not meaningful, we apply the same principle: augment reconstruction error with a structural consistency measure to obtain richer and more discriminative anomaly scores. Concretely, we use standard MSE for pointwise deviations and \emph{Reconstruction Rank-Order Consistency} \citep{singh2008order} to evaluate whether pairwise feature orderings are preserved between \(x\) and \(\hat{x}\). MSE reflects absolute reconstruction fidelity, while rank consistency exposes structural misalignments that purely distance-based scores may miss.

Nonetheless, it is important to emphasize that we do not claim these metrics are optimal for all datasets. Rather, the choice of scores should be guided by the types of shifts most relevant to a given application. For our considered datasets, metrics spanning multiple representational levels are particularly well-motivated.

\section{EXPERIMENTS}

In this section, we evaluate DISC on image and tabular benchmarks. We begin by detailing the datasets, models, and evaluation protocols, then present and discuss empirical results, examining standard ID–vs–OOD detection as well as OOD-type separability.
\begin{table*}[ht]
\centering
\begingroup
\setlength{\tabcolsep}{3.5pt}
\renewcommand{\arraystretch}{0.96}
\footnotesize
\sisetup{detect-weight=true,detect-inline-weight=math}

\caption{%
Average performance in the standard ID–vs–OOD setting and the proposed multi‐OOD settings for two in‐distribution datasets (CIFAR-10 and ImageNet).
\textbf{Avg AUROC}: area under the ROC curve for classic OOD detection.
\textbf{Clust. Acc}: K-means accuracy in the multi‐OOD setting.
\textbf{Sup. Acc}: supervised MLP accuracy in the multi‐OOD setting.
For \textbf{ID = ImageNet}, OOD sets include ImageNet-A/O/C, CIFAR-10, MNIST.
For \textbf{ID = CIFAR-10}, OOD sets include CIFAR-10-FGSM, horizontal flips, 20\% black-pixel occlusion, Butterflies, MNIST.
\textbf{‘All detectors’} denotes concatenation of all baseline scalar scores into one feature vector with the same clustering/MLP protocol as DISC.
Values are \textbf{mean $\pm$ std} over 5 seeds; per-dataset scores are in the Appendix.
}
\label{tab:ood_performance}

\resizebox{\textwidth}{!}{%
\begin{tabular}{%
    l
  | S[separate-uncertainty = true, table-format=1.4(2)]
    S[separate-uncertainty = true, table-format=1.4(2)]
    S[separate-uncertainty = true, table-format=1.4(2)]
  | S[separate-uncertainty = true, table-format=1.4(2)]
    S[separate-uncertainty = true, table-format=1.4(2)]
    S[separate-uncertainty = true, table-format=1.4(2)]
  }
\toprule
\multirow{2}{*}{\textbf{Detector}}
  & \multicolumn{3}{c|}{\textbf{ID: CIFAR10}}
  & \multicolumn{3}{c}{\textbf{ID: ImageNet}} \\
\cmidrule(lr){2-4} \cmidrule(lr){5-7}
  & \textbf{AUROC}
  & \textbf{Clust. Acc}
  & \textbf{Sup. Acc}
  & \textbf{AUROC}
  & \textbf{Clust. Acc}
  & \textbf{Sup. Acc} \\
\midrule
EnergyBased
  & 0.8187 \pm 0.0123 & 0.4164 \pm 0.0201 & 0.4309 \pm 0.0195
  & 0.5059 \pm 0.0104 & 0.2332 \pm 0.0137 & 0.2348 \pm 0.0142 \\
Entropy
  & 0.8291 \pm 0.0118 & 0.3690 \pm 0.0186 & 0.3832 \pm 0.0174
  & 0.5057 \pm 0.0107 & 0.2036 \pm 0.0128 & 0.2080 \pm 0.0131 \\
MCD
  & 0.8243 \pm 0.0121 & 0.3707 \pm 0.0193 & 0.3848 \pm 0.0179
  & 0.4814 \pm 0.0112 & 0.2075 \pm 0.0122 & 0.2113 \pm 0.0125 \\
MSP
  & 0.8257 \pm 0.0120 & 0.3553 \pm 0.0178 & 0.3755 \pm 0.0170
  & 0.5024 \pm 0.0106 & 0.1919 \pm 0.0117 & 0.2020 \pm 0.0120 \\
Mahalanobis
  & \bfseries 0.8415 \pm 0.011 & 0.4935 \pm 0.0214 & 0.4959 \pm 0.0205
  & 0.6403 \pm 0.0127 & 0.1869 \pm 0.0112 & 0.1898 \pm 0.0116 \\
Maha+ODIN
  & 0.8381 \pm 0.0111 & 0.4860 \pm 0.0221 & 0.5285 \pm 0.0209
  & 0.6530 \pm 0.0130 & 0.1829 \pm 0.0109 & 0.1834 \pm 0.0110 \\
MaxLogit
  & 0.8187 \pm 0.0125 & 0.4185 \pm 0.0192 & 0.4250 \pm 0.0188
  & 0.5059 \pm 0.0101 & 0.2305 \pm 0.0132 & 0.2302 \pm 0.0134 \\
ODIN
  & 0.7968 \pm 0.0130 & 0.3613 \pm 0.0181 & 0.4339 \pm 0.0190
  & 0.5328 \pm 0.0109 & 0.1929 \pm 0.0119 & 0.2184 \pm 0.0126 \\
ReAct
  & 0.7321 \pm 0.0142 & 0.3238 \pm 0.0167 & 0.3207 \pm 0.0171
  & 0.4784 \pm 0.0116 & 0.1763 \pm 0.0112 & 0.1743 \pm 0.0114 \\
ViM
  & 0.8020 \pm 0.0128 & 0.4250 \pm 0.0187 & 0.4408 \pm 0.0181
  & 0.5452 \pm 0.0112 & 0.2041 \pm 0.0123 & 0.2046 \pm 0.0125 \\
\midrule
\textbf{All detectors}
  &  \multicolumn{1}{c}{\text{--}} & 0.4907 \pm 0.0210 & 0.7084 \pm 0.0184
  &  \multicolumn{1}{c}{\text{--}} & 0.2343 \pm 0.0130 & 0.3374 \pm 0.0152 \\
\midrule
\textbf{DDPM \citep{graham2023denoising}}
  & 0.7529 \pm 0.0126 & 0.3967 \pm 0.0189 & 0.3816 \pm 0.0176
  & 0.6502 \pm 0.0121 & 0.3275 \pm 0.0165 & 0.3905 \pm 0.0172 \\
\textbf{DISC (ours)}
  & 0.8261 \pm 0.0108 & \bfseries 0.5120 \pm 0.0110  & \bfseries 0.7131 \pm 0.0180
  & \bfseries 0.7404 \pm 0.0119 & \bfseries 0.4776 \pm 0.0194 & \bfseries 0.6203 \pm 0.0163 \\
\bottomrule
\end{tabular}
}
\endgroup
\end{table*}

\subsection{Experimental Set-Up}
We adopt the Variance Preserving formulation (as in DDPMs \citep{ho2020denoising}), which interpolates between the data distribution and a fixed-variance Gaussian.  By concatenating the reconstruction‐error and histogram‐based statistics across all noise levels, we obtain a high-dimensional feature vector \(s(x)\). We evaluate this embedding under three different settings:  \\
\begin{enumerate}
    \item \textbf{Unsupervised anomaly detection}: using an IForest \citep{liu2008isolation} fitted exclusively on in-distribution feature vectors to assign anomaly scores.
    \item \textbf{Unsupervised clustering}: using K-means with the number of clusters fixed to the number of OOD categories, assessing whether distinct shifts can be separated without labels.
    \item \textbf{Supervised classification}:  using a two-layer MLP trained on held-out in-distribution and OOD samples from different categories of Distribution shifts, measuring performance when limited supervision is available.
\end{enumerate}
This evaluation framework probes the richness and disentanglement of the multi-statistic embedding \(s(x)\): the first scenario aligns with the standard OOD detection setting, while the latter two allow us to investigate whether the representation supports finer-grained discrimination of diverse distributional shifts. Our aim here is not merely to maximize raw performance, but to characterize the type of information encoded in \(s(x)\) under different conditions. To control for the dimensionality advantage of DISC, we also form a feature-rich baseline (‘All detectors’). This concatenates the outputs of several strong scalar OOD scores into a single feature vector. We then apply identical protocols on this concatenated vector.
 
\subsection{Image Analysis}
We empirically corroborate our hypothesis on the usefulness of a multi-statistic design with experiments on ImageNet and CIFAR-10. Recall that our central claim is that time-expanded reconstructions, when paired with complementary metrics, provide feature representations that not only achieve competitive unsupervised OOD detection, but also enable better differentiation between distinct types of distributional shift. To test this, we compare various common models as well as embeddings derived from single metrics (e.g., MSE alone) against the full multi-statistic embedding that integrates pixel-level, perceptual, structural, and frequency descriptors across time-steps. We emphasize that Avg.~AUROC is included to maintain comparability with the standard binary OOD literature, whereas Clust.~Acc. and Sup.~Acc. directly evaluate the finer-grained OOD-type characterization capability that DISC is designed to provide.

The results in Table~1 support our hypothesis. On ImageNet, a DDPM using only MSE reconstruction error already matches or exceeds several baselines in Avg.~AUROC and provides clear gains in multi-OOD discrimination. Extending this to the full DISC embedding yields further improvements across all three evaluation settings, with Avg.~AUROC increasing to 0.7404, K-means accuracy to 0.4776, and supervised MLP accuracy to 0.6203. Results on CIFAR-10 show the same overall pattern, with DISC consistently achieving the strongest multi-OOD classification performance. Additional ablations on the contribution of the full metric suite and on the role of diffusion timestep subsets are deferred to the Appendix.


\subsection{Tabular Data Analysis}
\begin{center}
\includegraphics[width=\linewidth]{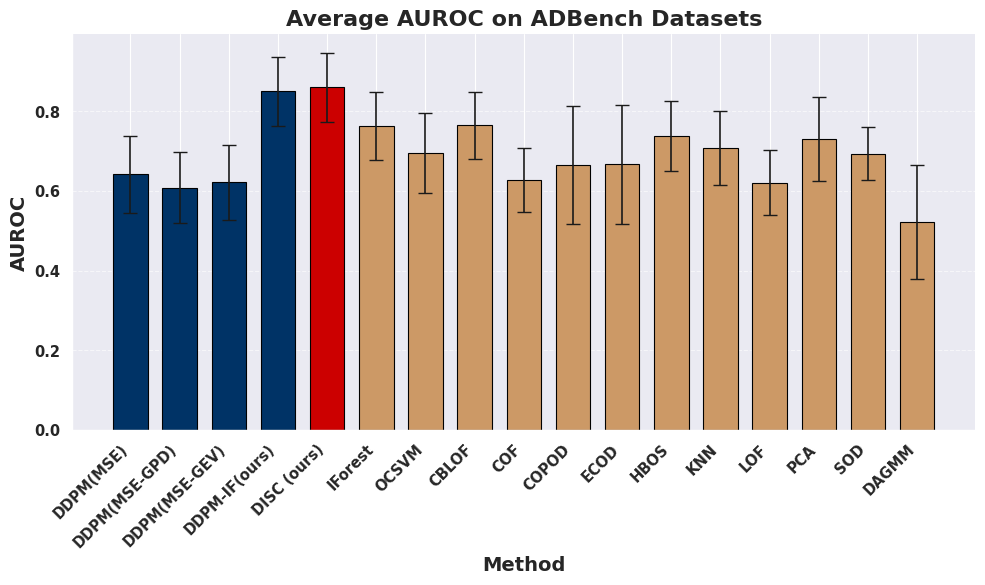}
\captionof{figure}{Mean AUC-ROC scores on ADBench for unsupervised anomaly detection. Our method, DISC (red), uses an Isolation Forest on combined rank-consistency and reconstruction error scores. It is compared to diffusion-based methods (blue) and tabular detectors (dark yellow). The blue bars show that for diffusion-based approaches, the thresholding choice is critical, with IForest outperforming other schemes.}
\label{tab_fig}
\end{center}

Next, we empirically investigate whether the principles of our framework extend beyond vision to the tabular domain. 
Experiments are conducted on a subset of datasets from the ADBench benchmark \citep{han2022adbench}, with sufficient training samples for fairer comparisons.

Again, the results provide strong evidence in favor of our hypothesis. DISC achieves an average AUROC of 0.86 (Figure~\ref{tab_fig}), surpassing the strongest comparable reconstruction-based baseline (DDPM with MSE-iForest, 0.84) and significantly outperforming other methods. The consistent gain over the MSE-only approach underscores the value of structural consistency: distributional shifts often disrupt feature relationships without large absolute errors, which further illustrates the importance of multi-statistic time-expanded reconstruction as a base framework.

\section{CONCLUSION}
We presented DISC: diffusion-based statistical characterization. DISC draws motivation from the observation that scalar detectors collapse heterogeneous shifts. As an alternative, DISC thus extracts complementary signals from diffusion trajectories to separate OOD families that classic scores tend to conflate. Across image and tabular benchmark corroboration, DISC yields consistent gains in standard ID–vs–OOD detection and, crucially, is able to classify \emph{which} OOD family a sample belongs to. A natural trade-off of DISC is computational cost. Because it relies on denoising evaluations across multiple noise levels and on a richer metric suite, its inference cost is higher than that of standard single-pass scalar detectors. In this sense, DISC is better viewed alongside higher-cost uncertainty estimators such as Deep Ensembles or Monte Carlo Dropout, which also trade additional inference computation for improved uncertainty signals. Our results suggest that this added overhead is justified when the goal extends beyond binary rejection and toward OOD type characterization: unlike standard detectors, DISC provides a representation that supports both competitive binary detection and substantially stronger separation between distinct OOD families.

Overall, our work  challenges the prevailing paradigms of OOD detection, which we demonstrate is fundamentally limited both empirically and theoretically. We hope that our work inspires a move beyond the simple ID–vs–OOD setting toward richer open-world evaluations and enable systems that reason about the \emph{nature} of novelty.

\section*{Acknowledgments}
This work was supported by the KIBA Project (Künstliche Intelligenz und diskrete Beladeoptimierungsmodelle zur Auslastungssteigerung im Kombinierten Verkehr - KIBA) under reference number 45KI16E051, funded by the German Ministry of Transportation.

 \bibliography{references}

\onecolumn
\appendix

\aistatstitle{Beyond Binary Out of Distribution Detection: Characterizing Distributional Shifts with Multi-Statistic Diffusion Trajectories: \\
Supplementary Materials}

\section{Impossibility of \(\phi\)-based OOD Tests: Indistinguishable Alternatives}

\subsection*{ Proposition}
Let \(P\) be a probability measure on a sample space \(\mathcal X\) and let 
\(\phi_{p}\colon\mathcal X\to\mathbb R\) be any (measurable) test statistic.  
Assume there exists a measurable set 
\[
  \Phi_{0}\subseteq\mathrm{Range}(\phi_{p})
  \quad\text{with}\quad
  P\!\bigl(\phi_{p}(X)\in\Phi_{0}\bigr)\;>\;0
\]
such that, for almost every \(\phi\in\Phi_{0}\), the conditional law 
\(P(\,\cdot\mid\phi_{p}(X)=\phi)\) is \emph{non-degenerate} (i.e.\ not a point mass).
Then there exist two distributions \(Q_{1}\) and \(Q_{2}\) satisfying
\[
  Q_{1}\neq P,\qquad Q_{2}\neq P,\qquad Q_{1}\neq Q_{2},
  \qquad\text{but}\qquad
  \phi_{p}(X)\sim Q_{1}= \phi_{p}(X)\sim Q_{2}= \phi_{p}(X)\sim P .
\]

All three distributions induce the same marginal law on the statistic $\phi_p(X)$.
As a result, any hypothesis test that depends only on \(\phi_{p}(X)\) attains
power equal to its false-positive rate against both \(Q_{1}\) and \(Q_{2}\).

\subsection*{Proof}
\noindent\textbf{Proof Sketch.}
\noindent We construct two distributions, \(Q_1\) and \(Q_2\), that are distinct from \(P\) and each other, yet engineered to produce the exact same marginal distribution for the statistic \(\phi_p(X)\). The strategy is to identify a set of statistic values, \(\Phi\), where the conditional distribution \(P(X \mid \phi_p(X) = \phi)\) is not concentrated on a single point. For each \(\phi \in \Phi\), this non-degeneracy allows us to find at least two disjoint subsets of the outcome space, \(A_\phi\) and \(B_\phi\), which both have positive conditional probability. We then define new conditional distributions, \(q_1(x \mid \phi)\) and \(q_2(x \mid \phi)\). For \(\phi \in \Phi\), these new distributions slightly re-weight the probability mass between the sets \(A_\phi\) and \(B_\phi\). For any \(\phi \notin \Phi\), we leave the conditional distribution unchanged \(\bigl(q_i(x \mid \phi) = p(x \mid \phi)\bigr)\). Finally, we construct the full distributions \(Q_1\) and \(Q_2\) by combining these new conditional distributions with the original marginal density of \(\phi_p(X)\). This procedure guarantees that the marginal law of \(\phi_p(X)\) is identical under \(P\), \(Q_1\), and \(Q_2\), while the joint distributions are provably different. Any test that depends only on $\phi_{p}(X)$ therefore attains identical
rejection probabilities under all three measures, so its power equals its
false-positive rate.


\textbf{Non-Degeneracy Assumption.}
By assumption, the family of conditional distributions \( p(x \mid \phi_p(x)) \) is non-degenerate on a set of statistic-values of positive measure.
Formally, there exists a measurable set
\[
  \Phi\subseteq\mathrm{Range}(\phi_{p})
  \quad\text{with}\quad
  \mu(\Phi)\;=\;P\!\bigl(\phi_{p}(X)\in\Phi\bigr)\;>\;0,
\]
such that for \(\mu\)-almost every \(\phi\in\Phi\) one can find a measurable
subset
\[
  A_{\phi}\subseteq\operatorname{supp}p(\,\cdot\mid\phi)
  \quad\text{satisfying}\quad
  0<P\!\bigl(X\in A_{\phi}\mid\phi_{p}(X)=\phi\bigr)<1.
\]

\medskip
For each \(\phi\in\Phi\) define the level set
\[
  S_{\phi}\;=\;\{\,x\in\mathcal X:\phi_{p}(x)=\phi\,\}
\]
Because the conditional law is non-degenerate, the subset \(A_{\phi}\)
is a \emph{proper} part of \(S_{\phi}\).
Set
\[
  B_{\phi}\;=\;S_{\phi}\setminus A_{\phi}.
\]
Then, for \(\mu\)-almost every \ \(\phi\in\Phi\),
\[
  A_{\phi}\cap B_{\phi}=\varnothing,\qquad
  A_{\phi}\cup B_{\phi}=S_{\phi},\qquad
  0<P(A_{\phi}\mid\phi)<1,\qquad
  0<P(B_{\phi}\mid\phi)<1.
\]

\paragraph{Definition of the Test Function $f_{p}(x)$}  
We now construct a measurable function \(f_{p}(x)\) that separates \(P\) and \(Q\) under the law of expectation.  
Let \(g(x)\) be any measurable function with \(\mathbb{E}_{p(x\mid\phi)}[\,|g(x)|\,]<\infty\) for all \(\phi\notin\Phi\) .  Fix the partition \(A_{\phi},B_{\phi}\subseteq \operatorname{supp}(p(x\mid\phi))\) for each \(\phi\in\Phi\) as above.  Then define
\begin{equation}
  f_{p}(x)
  \;=\;
  \begin{cases}
    1, 
      & \phi_{p}(x)\in \Phi
        \;\text{and}\;
        x\in A_{\phi_{p}(x)}, \\[6pt]
    \frac{1}{2}, 
      & \phi_{p}(x)\in \Phi
        \;\text{and}\;
        x\in B_{\phi_{p}(x)}, \\[6pt]
    g(x),
      & \phi_{p}(x)\notin \Phi.
  \end{cases}
  \label{eq:fp_extended}
\end{equation}

\begin{itemize}[leftmargin=*]
  \item If \(\phi_{p}(x)\in \Phi\), then \(f_{p}(x)\) takes value \(1\) on the “upper” set \(A_{\phi}\) and ($\frac{1}{2}$) on the complementary piece \(B_{\phi}\).  
  \item If \(\phi_{p}(x)\notin \Phi\), we revert to \(g(x)\), ensuring \(f_{p}\) remains integrable and well‐behaved off \(\Phi\).  
   
\end{itemize}

\paragraph{Construction of the Alternative Conditional Distributions \(q_{1}(x \mid \phi_{p}(x))\) and \(q_{2}(x \mid \phi_{p}(x))\).}

Fix a constant \(0<\lambda<1\).
For every statistic value \(\phi\) in the \emph{essential range} of
\(\phi_{p}\) we define new conditional densities:
(The selector \(\phi\mapsto A_{\phi}\) is measurable, so the map
\((\phi,x)\mapsto q_{i}(x\mid\phi)\) is jointly measurable.)

\[
\begin{aligned}
  q_{1}\bigl(x \mid \phi_{p}(x)=\phi\bigr)
  &= 
  \begin{cases}
    \displaystyle
    \frac{1}{C^{(1)}_{\phi}}
    \Bigl[
      \lambda\,p\bigl(x \mid \phi\bigr)\,\mathbbm{1}\{\,x\in A_{\phi}\}
      \;+\;
      p\bigl(x \mid \phi\bigr)\,\mathbbm{1}\{\,x\in B_{\phi}\}
    \Bigr], 
    & \phi \in \Phi, 
    \\[1em]
    p\bigl(x \mid \phi\bigr), 
    & \phi \notin \Phi,
  \end{cases}
  \\[1.5em]
  q_{2}\bigl(x \mid \phi_{p}(x)=\phi\bigr)
  &= 
  \begin{cases}
    \displaystyle
    \frac{1}{C^{(2)}_{\phi}}
    \Bigl[
      p\bigl(x \mid \phi\bigr)\,\mathbbm{1}\{\,x\in A_{\phi}\}
      \;+\;
      \lambda\,p\bigl(x \mid \phi\bigr)\,\mathbbm{1}\{\,x\in B_{\phi}\}
    \Bigr], 
    & \phi \in \Phi, 
    \\[1em]
    p\bigl(x \mid \phi\bigr), 
    & \phi \notin \Phi,
  \end{cases}
\end{aligned}
\]
where the normalization constants for each \(\phi\in\Phi\) are
\[
\begin{aligned}
  C^{(1)}_{\phi}
  &= \int_{A_{\phi}} \lambda\,p(x\mid\phi)\,dx
    \;+\;\int_{B_{\phi}} p(x\mid\phi)\,dx
  = \lambda\,P_{p(x\mid\phi)}(A_{\phi})
    + P_{p(x\mid\phi)}(B_{\phi}),
  \\[6pt]
  C^{(2)}_{\phi}
  &= \int_{A_{\phi}} p(x\mid\phi)\,dx
    \;+\;\int_{B_{\phi}} \lambda\,p(x\mid\phi)\,dx
  = P_{p(x\mid\phi)}(A_{\phi})
    + \lambda\,P_{p(x\mid\phi)}(B_{\phi}).
\end{aligned}
\]

\begin{itemize}[leftmargin=*]
  \item For \(\phi\notin\Phi\) no change is made:
      \(q_{i}(\,\cdot\mid\phi)=p(\,\cdot\mid\phi)\) for \(i=1,2\).
  \item For \(\phi\in\Phi\):
    \begin{itemize}
      \item \(q_{1}\) “down‐weights” the mass on \(A_{\phi}\) by \(\lambda\), leaving \(B_{\phi}\) unchanged, then renormalizes by \(C^{(1)}_{\phi}\).  
      \item \(q_{2}\) “down‐weights” the mass on \(B_{\phi}\) by \(\lambda\), leaving \(A_{\phi}\) unchanged, then renormalizes by \(C^{(2)}_{\phi}\).  
    \end{itemize}
  \item Since \(0<\lambda<1\) and \(0<P(A_{\phi}),P(B_{\phi})<1\), one checks
    \[
      0 < C^{(1)}_{\phi},\,C^{(2)}_{\phi} < 1,
      \quad
      \frac{\lambda}{C^{(1)}_{\phi}}<1,
      \quad
      \frac{\lambda}{C^{(2)}_{\phi}}<1.
    \]
 
\end{itemize}
\paragraph{Support Preservation.}
 Since $0<\lambda<1$ and $C^{(i)}_{\phi_{p}(x)}>0$, we have
\[
q_{i}(x\mid\phi_{p}(x))>0
\;\Longleftrightarrow\;
p(x\mid\phi_{p}(x))>0.
\]
Thus every $x$ with $p(x\mid\phi_{p}(x))>0$ also satisfies $q_{i}(x\mid\phi_{p}(x))>0$, and vice versa, proving the supports coincide. For each $i=1,2$ and every $\phi_{p}(x)$, 
\[
\mathrm{supp}\bigl(q_{i}(\,\cdot\mid\phi_{p}(x))\bigr)
\;=\;
\mathrm{supp}\bigl(p(\,\cdot\mid\phi_{p}(x))\bigr).
\]

The parameter \(\lambda\) controls the redistribution of mass between \(A_\phi\) and \(B_\phi\) within each level set without violating support preservation. If \(\lambda = 1\), then \(q_i(x \mid \phi)\) reduces to \(p(x \mid \phi)\), so \(Q_i = P\), violating the goal of constructing distinct alternatives. If \(\lambda = 0\), one region is entirely zeroed out, violating support preservation. Choosing \(0<\lambda<1\) yields two
valid, distinct alternatives \(Q_{1},Q_{2}\) that share the same
marginal distribution of \(\phi_{p}(X)\) as \(P\).

\paragraph{Defining the Full Distributions \(\boldsymbol{Q_{1}}\) and \(\boldsymbol{Q_{2}}\).}
Having specified the conditional probability kernels \(q_{i}(dx\mid\phi)\), we construct the full measures by combining them with the original marginal law of the statistic \(\phi_p(X)\).

Let \(p_{\phi}(d\phi)\) be the push-forward measure of \(P\) by the map \(\phi_p\). We define the joint measures on the product space as:
\[
 Q_{i}(dx,d\phi)\;=\;p_{\phi}(d\phi)\;q_{i}(dx\mid\phi),
 \qquad i=1,2.
\]
The marginal measure on \(\mathcal{X}\) is found by integrating over \(\phi\):
\[
 Q_i(dx) = \int_{\mathrm{Range}(\phi_p)} q_i(dx \mid \phi) \, p_\phi(d\phi).
\]
If \(p_\phi(\cdot)\) is the density for the measure \(p_\phi(d\phi)\), and \(q_i(\cdot \mid \phi)\) is the density for \(q_i(dx \mid \phi)\), then the density \(q_i(x)\) for the measure \(Q_i(dx)\) is given by:
\begin{equation}
 q_i(x) = p_\phi(\phi_p(x)) \times q_i(x \mid \phi_p(x)).
 \label{eq:qi_full_density}
\end{equation}

\begin{itemize}[leftmargin=*]
\item Because \(\int q_{i}(dx\mid\phi)=1\) for every \(\phi\), each \(Q_i\) is a valid probability measure.
\item By construction, the push-forward of each \(Q_i\) by \(\phi_p\) is \(p_\phi(d\phi)\). Therefore, \(P, Q_1,\) and \(Q_2\) all share the same marginal law for the statistic.
\end{itemize}

\paragraph{Marginal matching of \(\phi_{p}(X)\).}
Let \(p_{\phi}(d\phi)=P\!\bigl(\phi_{p}(X)\in d\phi\bigr)\) be the push-forward
measure of \(P\).
For every measurable set \(B\subseteq\mathrm{Range}(\phi_{p})\),
\[
  Q_{i}\bigl(\phi_{p}(X)\in B\bigr)
  =\int_{B}\!\int q_{i}(dx\mid\phi)\,p_{\phi}(d\phi)
  =\int_{B} p_{\phi}(d\phi)
  =P\bigl(\phi_{p}(X)\in B\bigr),\qquad i=1,2,
\]
where the middle equality uses
\(\int q_{i}(dx\mid\phi)=1\) for all \(\phi\)
(joint measurability established earlier).
Hence
\[
  Q_{1}\circ\phi_{p}^{-1}
  =Q_{2}\circ\phi_{p}^{-1}
  =P\circ\phi_{p}^{-1},
\]
i.e.\ the statistic \(\phi_{p}(X)\) has the same distribution
under \(P, Q_{1}, Q_{2}\).

\paragraph{Verifying \(\boldsymbol{Q_{i}\neq P}\) and \(\boldsymbol{Q_{1}\neq Q_{2}}\).} Our goal is now to show that the expectations of the test function \( f_p \) under \( P \) and under the alternative distributions \( Q_1 \) and \( Q_2 \) differ, i.e.,
\[
\mathbb{E}_{P}[f_p(X)] - \mathbb{E}_{Q_i}[f_p(X)] > 0 \quad \text{for } i = 1, 2,
\]
using the law of total expectation.

Write \(p_{\phi}(d\phi)=P\bigl(\phi_{p}(X)\in d\phi\bigr)\) for the
push-forward measure of \(P\). For every \(\phi\in\Phi\) for which the partition is defined, set
\(a_\phi:=P(A_\phi\mid\phi)\) and \(b_\phi:=P(B_\phi\mid\phi)=1-a_\phi\).
Recall that the test function is \(f_{p}(x)=1\) on \(A_\phi\) and \(f_{p}(x)=\frac{1}{2}\) on \(B_\phi\).

\smallskip\noindent

For \(\phi\in\Phi\), the conditional expectations are:
\[
 \mathbb{E}_{P}\!\bigl[f_{p}(X)\mid\phi\bigr]=a_\phi+\tfrac{1}{2}\,b_\phi,
 \qquad
 \mathbb{E}_{Q_{1}}\!\bigl[f_{p}(X)\mid\phi\bigr]
   =\frac{\lambda a_\phi+\tfrac{1}{2}\,b_\phi}{\lambda a_\phi+b_\phi},
 \qquad
 \mathbb{E}_{Q_{2}}\!\bigl[f_{p}(X)\mid\phi\bigr]
   =\frac{a_\phi+\tfrac{1}{2}\,\lambda b_\phi}{a_\phi+\lambda b_\phi}.
\]
For \(\phi\notin\Phi\), all three conditional expectations are equal, so these terms cancel when computing differences of the full expectations.

\smallskip\noindent
For any \(0<\lambda<1\) and \(0<a_\phi,b_\phi<1\), the difference in conditional expectations is:
\[
 \Delta_{1}(\phi):=
 \mathbb{E}_{P}[f_p\mid\phi]-\mathbb{E}_{Q_{1}}[f_p\mid\phi]
 =\frac{0.5(1-\lambda)\,a_\phi\,b_\phi}{\lambda a_\phi+b_\phi}>0.
\]
Since \(\Delta_1(\phi) > 0\) on the set \(\Phi\) which has positive measure, the integral is strictly positive:
\[
 \mathbb{E}_{P}[f_{p}(X)]-\mathbb{E}_{Q_{1}}[f_{p}(X)]
 =\int_{\Phi}\!\Delta_{1}(\phi)\,p_{\phi}(d\phi)\;>\;0,
\]
so \(Q_{1}\neq P\).

\smallskip\noindent

Similarly for \(Q_{2}\):
\[
 \Delta_{2}(\phi) := \mathbb{E}_{P}[f_p\mid\phi]-\mathbb{E}_{Q_{2}}[f_p\mid\phi]
 = \frac{0.5(1-\lambda)\,a_\phi b_\phi}{a_\phi+\lambda b_\phi}>0.
\]
Therefore \(\mathbb{E}_{P}[f_p]-\mathbb{E}_{Q_{2}}[f_p]>0\), and hence \(Q_{2}\neq P\).

\smallskip\noindent
For \(\phi\in\Phi\), the difference in conditional expectations is:
\[
 \mathbb{E}_{Q_{1}}[f_p\mid\phi]-\mathbb{E}_{Q_{2}}[f_p\mid\phi]
 =\frac{0.5\,a_\phi b_\phi(\lambda^2-1)}{(\lambda a_\phi+b_\phi)(a_\phi+\lambda b_\phi)}.
\]
Since \(0<\lambda<1\), the term \((\lambda^2-1)\) is strictly negative. Because \(a_\phi, b_\phi\) are strictly positive on \(\Phi\), the entire expression is non-zero. The integral of this difference over \(\Phi\) is therefore non-zero, so
\(\mathbb{E}_{Q_{1}}[f_p(X)]\neq\mathbb{E}_{Q_{2}}[f_p(X)]\), which proves \(Q_{1}\neq Q_{2}\).

\medskip\noindent
Hence \(Q_{1},Q_{2}\) are both distinct from \(P\) and from each other.

\paragraph{Indistinguishability by \(\phi_{p}\)-based Tests.}
Any test that depends solely on the statistic \(\phi_{p}(x)\) can be written as
\[
  \text{Reject }H_{0}
  \quad\Longleftrightarrow\quad
  \phi_{p}(x)\in R
  \quad\text{for some }R\subseteq \mathrm{Range}(\phi_{p}).
\]
Under \(P\), the false‐positive rate (size) of this test is
\[
  \mathrm{FPR}
  = P\bigl(\phi_{p}(X)\in R\bigr).
\]
Under either alternative \(Q_{i}\) (\(i=1,2\)), the test’s power is
\[
  \mathrm{Power}_{i}
  = Q_{i}\bigl(\phi_{p}(X)\in R\bigr).
\]
But since \(Q_{i}(\phi_{p}(X)\in R)=P(\phi_{p}(X)\in R)\) for both \(i\), we have
\[
  \mathrm{Power}_{i} \;=\;\mathrm{FPR}
  \quad (\text{for }i=1,2).
\]
Hence every \(\phi_{p}\)-based test has power equal to size under \(Q_{1}\) and \(Q_{2}\). 
Outline for Appendix:
1. Implementation Details 
2. Additional Results More Detailed Results Dataset wise
3. Additional Detailed Results for the baselines showing how they struggle
4. The Results showcasing per Operator Performance
5.Tabular Data detailed Results
6. Tabular Data distinction between different types.

\section{Extended Background}
\subsection{Out-of-Distribution Detection Score Computation}

Out-of-distribution (OOD) detection methods assign each input \(x\) an OOD score \(o(x)\) that reflects how likely \(x\) lies outside the training distribution. Previous works have attempted to categorize these methods in various ways; however, most modern approaches leverage ideas from multiple paradigms to compute OOD scores. In the case of generative models, one can compute an OOD score either from the estimated density or from a reconstruction loss.

\paragraph{Density-Based Methods}  
Density-based methods learn an explicit generative model \(p(x)\) over in-distribution data and use the negative log-likelihood as the OOD score:
\[
o(x) \;=\; -\log p(x).
\]
Under this formulation, inputs falling into low-density regions—i.e., those that yield small values of \(p(x)\)—incur large values of \(o(x)\) \citep{nalisnick2018deep}. For instance, PixelCNN++ is an autoregressive model that estimates image likelihoods using convolutional networks and computes \(p(x)\) exactly via a tractable factorization \citep{salimans2017pixelcnn}. Likewise, Glow employs invertible flow layers that permit exact likelihood evaluation through a sequence of bijective transformations \citep{kingma2018glow}. In both cases, test inputs that lie far from the training distribution receive high OOD scores via \(-\log p(x)\).

\paragraph{Reconstruction-Based Methods}  
Reconstruction-based methods rest on the assumption that in-distribution samples lie on or near a low-dimensional manifold embedded in the input space. One trains a reconstruction model \(r(x)\) so that, for an in-distribution sample, \(r(x)\approx x\). The OOD score is then taken to be the reconstruction error:
\[
o(x) \;=\; \bigl\lVert\,x \;-\; r(x)\bigr\rVert_{2}.
\]
Since OOD inputs do not conform to the learned manifold, they tend to yield large reconstruction errors. Common instantiations include variational autoencoders trained with a pixel-wise \(L_{2}\) reconstruction loss, as well as diffusion models that are optimized to minimize the same loss on training data \citep{oko2023diffusion,livernoche2023diffusion}. In practice, an input \(x\) whose reconstruction \(r(x)\) deviates substantially from \(x\) is flagged as likely OOD.

\paragraph{Discriminative Models and Calibration}  
Discriminative models are typically trained on a supervised classification task and leverage outputs (logits) or internal features (e.g., from the penultimate layer) for a given input \(x\) for OOD Detection. Because such models often yield poorly calibrated confidences, especially on inputs outside the training distribution, many approaches first focus on calibration before computing a post-hoc score. One prominent calibration technique is \emph{Outlier Exposure}, which augments the training process with a diverse set of outlier examples and penalizes high confidence on these samples \citep{hendrycks2018deep}. Formally, let \(p_{\theta}(y \mid x)\) denote the softmax probability of class \(y\). During training, an auxiliary loss term encourages \(p_{\theta}(y \mid x_{\text{out}})\) to remain near uniform for outlier inputs \(x_{\text{out}}\), thereby reducing overconfidence on novel examples. Another calibration strategy is \emph{ODIN}, which applies temperature scaling to the logits and adds a small adversarial perturbation to the input; this softens the softmax outputs and amplifies the gap between in-distribution (ID) and OOD confidences. 

\paragraph{Post-Hoc Logit-Based Scores}  
Once a discriminative model \(f\colon x \mapsto \{f_{c}(x)\}_{c=1}^{C}\) is trained (and potentially calibrated), one can derive OOD scores directly from its logits \(\{f_{c}(x)\}\). The simplest baseline is the \emph{Maximum Softmax Probability} (MSP), defined as
\[
o_{\mathrm{MSP}}(x) \;=\; -\max_{c}\;\softmax\bigl(f(x)\bigr)_{c}.
\]
Although intuitive, MSP often remains overconfident on novel inputs, yielding poor separation between ID and OOD samples \citep{hendrycks2016baseline}. A more informative metric from information theory is the \emph{predictive entropy}:
\[
o_{\mathrm{ent}}(x) \;=\; - \sum_{c=1}^{C} p(c \mid x)\,\log p(c \mid x),\quad p(c \mid x)=\softmax\bigl(f(x)\bigr)_{c}.
\]
Higher entropy indicates greater uncertainty, and thus a higher likelihood of being OOD. To produce better-calibrated uncertainty estimates, one can employ \emph{Monte Carlo Dropout}—which aggregates stochastic predictions under dropout at test time \citep{gal2016dropout}—or \emph{Deep Ensembles}, which average the outputs of multiple independently trained models \citep{lakshminarayanan2017simple}. Both techniques approximate Bayesian inference and tend to assign higher entropy to inputs that deviate from the training distribution.

The \emph{Energy Score} provides another logit-based measure by mapping logits to a scalar via a log-sum-exp operation:
\[
o_{E}(x) \;=\; -\log \sum_{c=1}^{C} \exp\bigl(f_{c}(x)\bigr).
\]
Since this score does not normalize the logits into a simplex, it mitigates the overconfidence issues of softmax; in practice, in-distribution samples produce strongly negative energy values, while OOD samples yield higher (less negative) energies \citep{liu2020energy}.

\paragraph{Post-Hoc Feature-Based Scores}  
Instead of relying on the logits, feature-based approaches compute OOD scores using internal representations—typically the output of the penultimate layer, denoted \(z(x)\). A widely used method is the \emph{Mahalanobis Distance} score: one fits a class-conditional Gaussian mixture model over the features, estimating a mean vector \(\mu_{c}\) for each class \(c\) and a (shared) covariance matrix \(\Sigma\). The Mahalanobis score is then given by
\[
o_{\mathrm{Maha}}(x) \;=\; \min_{c}\;(z(x)-\mu_{c})^{\top}\,\Sigma^{-1}\,(z(x)-\mu_{c}).
\]
Inputs whose features lie far from all class-conditional Gaussians receive large scores and are flagged as OOD \citep{lee2018simple,ren2021simple}. Alternatively, \emph{Deep \(k\)-Nearest Neighbors} (Deep \(k\)-NN) computes the distance to the nearest neighbors among training embeddings, effectively measuring how isolated \(z(x)\) is within the in-distribution feature manifold \citep{sun2022out}. Extreme Value Theory (EVT) can further refine thresholds on tail distances by fitting a distribution to the largest observed distances, thereby yielding more robust OOD detection \citep{ren2021simple}.

\paragraph{Gradient-Based Metrics}  
Gradient-based methods exploit the observation that OOD inputs, lying off the learned manifold, often induce larger gradients with respect to the input. One such metric is \emph{GradNorm}, which measures the norm of the input gradient of the loss at the predicted label \(\hat{y}\):
\[
o_{\mathrm{Grad}}(x) \;=\; \bigl\lVert \nabla_{x}\,\ell\bigl(\hat{y}\mid x\bigr)\bigr\rVert_{2}.
\]
Because OOD samples typically incur higher input gradients, a larger value of \(o_{\mathrm{Grad}}(x)\) signals a greater chance of being OOD \citep{huang2021importance}. More recent approaches fit the input’s Fisher information matrix to capture sensitivity peaks around OOD examples, thereby providing an alternative gradient-based OOD metric \citep{dauncey2024approximations}.

\paragraph{Inference-Modifying Post-Hoc Methods}  
Beyond computing a static score, certain methods modify the inference process itself to improve OOD detection without retraining the model. For example, one can prune or clip activations at test time—techniques such as \emph{ReAct}, which truncates excessively large ReLU activations \citep{sun2021react}; \emph{DICE}, which dynamically clips activations based on batch statistics \citep{sun2022dice}; and \emph{ASH}, which applies asymmetric activation clipping to preserve ID accuracy \citep{djurisic2022extremely}. After modifying activations, any of the aforementioned logit- or feature-based scores can be applied to achieve stronger separation between ID and OOD inputs.

It is important to note that many state-of-the-art methods weave together ideas from these different perspectives. For instance, one might compute a Mahalanobis score in the latent feature space of a generative model while using Monte Carlo Dropout to estimate predictive uncertainty, or measure input gradients of a diffusion-based generator. Consequently, the above categories—generative versus discriminative, density versus reconstruction, logit-based versus feature-based—should not be viewed as mutually exclusive; rather, they provide a taxonomy for understanding how various components may be combined to produce robust OOD detection.
\subsection{Score Based Generative Models}

Score-based generative models \citep{sohl2015deep, song2020score} learn to estimate the score function \(\nabla \log p(x)\) by training a collection of denoisers \(\{D_{\sigma}\}\) at varying noise levels. A \emph{forward diffusion} corrupts a clean sample \(x_0 \sim p(x)\) through a Markov chain of Gaussian perturbations:
\[
x_{\sigma_0} \;\longrightarrow\; x_{\sigma_1} \;\longrightarrow\; \cdots \;\longrightarrow\; x_{\sigma_\infty},
\]
with \(\sigma_0 < \sigma_1 < \cdots < \sigma_\infty\) and
\[
p(x_{\sigma} \mid x) = \mathcal{N}\bigl(x,\;\sigma^2 I\bigr), 
\quad
p_{\sigma}(x_{\sigma}) = \int \mathcal{G}_{\sigma}(x_{\sigma}-x)\,p(x)\,\mathrm{d}x.
\]

Tweedie’s formula \citep{efron2011tweedie} connects MMSE denoising to score estimation:
\begin{equation}
D_{\sigma}(x_{\sigma})
= \mathbb{E}[x \mid x_{\sigma}]
= x_{\sigma} + \sigma^2 \,\nabla \log p_{\sigma}(x_{\sigma}),
\end{equation}
so minimizing
\(\mathbb{E}\,\|x - D_{\sigma}(x_{\sigma})\|_2^2\)
recovers \(\nabla \log p_{\sigma}(x_{\sigma})\).

Based on the formulation of \citep{song2020score}, score-based models come in two main categories:

\begin{itemize}
  \item \textbf{Variance Preserving (VP)}  
    \[
      x_t = \sqrt{\alpha(t)}\,x_0 + \sqrt{1 - \alpha(t)}\,\epsilon,
      \quad \epsilon\sim\mathcal{N}(0,I),
    \]
    with \(\alpha(t)\in[0,1]\) decreasing, so \(\mathrm{Var}(x_t)\) remains bounded.
  \item \textbf{Variance Exploding (VE)}  
    \[
      x_t = x_0 + \sigma(t)\,\epsilon,
      \quad \epsilon\sim\mathcal{N}(0,I),
    \]
    where \(\sigma(t)\) increases, causing \(\mathrm{Var}(x_t)=\sigma^2(t)\to\infty\).
\end{itemize}

In practice, the denoiser \(D_{\sigma}\) is parameterized (e.g., by a neural network) and trained to minimize the mean squared error (MSE) between the clean sample \(x\) and its denoised estimate \(D_{\sigma}(x_{\sigma})\):
\begin{equation}
\mathrm{MSE}\bigl(D_{\sigma}\bigr) \;=\; \mathbb{E}_{x,\,x_{\sigma}} \bigl\lVert\,x - D_{\sigma}(x_{\sigma})\bigr\rVert_2^2.
\label{eq:mse_loss}
\end{equation}

A diffusion model thus comprises a collection of minimum‐mean‐square‐error denoisers \(\{D_{\sigma} : \sigma \in \sigma\}\). Each denoiser implicitly provides access to the corresponding score function \(\nabla \log p_{\sigma}(x_{\sigma})\). At sampling time, these learned score functions are employed to reverse the diffusion process and generate samples from the underlying clean distribution \(p(x)\) via a discretized reverse diffusion procedure.

\section{Related Work}

Out-of-distribution (OOD) detection methods are broadly categorized into generative and discriminative approaches. Generative models learn the in-distribution (ID) data distribution either through explicit density estimation (e.g., PixelCNN++ \citep{salimans2017pixelcnn}, Glow \citep{kingma2018glow}) or reconstruction-based techniques that identify OOD samples by their high reconstruction error (e.g., VAEs, diffusion models \citep{graham2023denoising}). However, the reliability of these density-based models has been challenged. Nalisnick et al. (2019) demonstrated that deep generative models can paradoxically assign a higher likelihood to OOD datasets than to the in-distribution data they were trained on, revealing a fundamental failure in their OOD detection capabilities. Further investigation by Zhang et al. (2021) attributed this failure to the models' reliance on simple background statistics or low-level features, which can cause them to assign high likelihood to OOD samples that share these trivial features with the training data \citep{zhang2021understanding}. 

Recent work has demonstrated that generative denoising processes are particularly effective OOD detectors \citep{graham2023denoising, livernoche2023diffusion}. For example, some methods use diffusion inpainting to reconstruct masked image regions, using the resulting error as an OOD score \citep{liu2023unsupervised}. Others have shown that a single, unconditional diffusion model can flag OOD instances by analyzing anomalies in its denoising trajectories \citep{heng2024out}. Another approach aggregates mismatches between noisy and denoised feature maps across diffusion steps, boosting detection sensitivity without requiring label supervision \citep{yang2024diffusion}.

Discriminative models, typically trained for classification, are repurposed for OOD detection using various scoring functions. These include logit-based approaches (e.g., MSP, energy scores \citep{hendrycks2016baseline, liu2020energy}), feature-based methods (e.g., Mahalanobis distance, Deep k-NN), gradient-based metrics \citep{huang2021importance}, and activation-based strategies (e.g., Tent \citep{rozsa2019improved}, DICE \citep{sun2022dice}). To mitigate classifier overconfidence on OOD inputs, calibration techniques like ODIN and Outlier Exposure are often employed \citep{hsu2020generalized, hendrycks2018deep}. While efficient at inference, the performance of these methods hinges on the classifier’s decision boundary and learned features, which are not always optimal for OOD detection \citep{li2025out}.

Modern approaches often combine generative and discriminative components to achieve robust performance. However, a fundamental trade-off exists between detecting "near" versus "far" OOD samples, suggesting that no single method excels at both \citep{guille2024expecting, zhang2021understanding}. Furthermore, many methods implicitly assume a unimodal, exponentially decaying score distribution. This assumption is adequate for binary ID–vs–OOD decisions but is ill-suited for complex, multi-modal OOD landscapes. The challenges are compounded by the finite-sample problem, especially in high-dimensional spaces where the curse of dimensionality makes it difficult to distinguish genuine distributional shifts from sampling artifacts \citep{ghosal2024overcome, aggarwal2001surprising}. Recent impossibility theorems have established that OOD detection is not generally PAC-learnable in distribution-free settings. Learnability is only possible under restrictive conditions on the hypothesis spaces and the relationship between ID and OOD data \citep{fang2022out, garov2025closer}.

\section{Implementation Details}


This section details datasets, models, hyperparameters, and evaluation protocols to reproduce our results.

\paragraph{Datasets and preprocessing.}
For images we use \textbf{CIFAR-10} and \textbf{ImageNet-1k} as in-distribution (ID) datasets. 
For CIFAR-10 we evaluate the following OOD families: \texttt{CIFAR-10-FGSM} (\(\epsilon=0.1\), steps \(=1\)), horizontal flips, \(90^\circ\) rotations, central black-pixel occlusion covering up to \(\approx 20\%\) of the image area, \texttt{Butterflies} (Smithsonian subset from HuggingFace), and \texttt{MNIST} (replicated to 3-channel RGB and resized). 
For ImageNet we use \texttt{ImageNet-A}, \texttt{ImageNet-O}, \texttt{ImageNet-C} ( no fixed severity index), \texttt{CIFAR-10}, and \texttt{MNIST} as OOD. 
CIFAR images are resized to \(32\times 32\) and normalized to \([{-}1,1]\) using per-channel mean/std \(\mu=(0.5,0.5,0.5), \sigma=(0.5,0.5,0.5)\); ImageNet images are resized to \(64\times 64\) and normalized with ImageNet statistics \(\mu=(0.485,0.456,0.406), \sigma=(0.229,0.224,0.225)\). When diffusion models are used, we (de)scale to \([{-}1,1]\) as required by the backbone. No test-time augmentation is used for OOD scoring. 

For tabular experiments we adopt the \texttt{ADBench} subset via our TabDDPM pipeline. Unless otherwise noted in per-dataset configs, numeric features are standardized (z-scoring) on the ID training split, categorical features use either \texttt{one-hot} or leave-one-out \texttt{counter} encoding, numeric missing values are imputed with the mean (or rows dropped where configured), categorical missing values use the most-frequent category, and we do not apply outlier clipping.

\paragraph{Backbones for baselines.}
For image-based baselines we use \texttt{WideResNet-28-10} pretrained on \texttt{CIFAR-10} (for CIFAR experiments) and \texttt{ResNet-50} pretrained for ImageNet experiments. These backbones are used as-is (no additional training/label smoothing). Representation-space methods (Mahalanobis, SHE, ViM, kNN when used, DICE/ASH/ReAct) operate on \emph{penultimate avg-pool} features. For tabular baselines we wire \texttt{DeepSVDD}, \texttt{ICL}, \texttt{DevNet}, and \texttt{SLAD}; model depth/width and optimizer hyperparameters follow the dataset-specific configs in our Code.

\paragraph{Diffusion backbone (DISC).}
We adopt two complementary VP–style DDPM backbones, both with U-Net denoisers, but used for different roles. For analyses on \(64\times64\) images we rely on the \emph{improved-diffusion} implementation, instantiated once with a cosine noise schedule and \(3000\) diffusion steps. The model predicts \(\epsilon\) without label conditionning, and uses timestep and learned-sigma rescaling. Its U-Net has \(128\) base channels with multipliers \([1,2,2,2]\), three residual blocks per stage, four attention heads, attention at resolutions \(16\) and \(8\), zero dropout, and scale–shift normalization. We train DDPM  with a linear \(\beta\) schedule (\(\beta_{\min}{=}10^{-4},\,\beta_{\max}{=}2{\times}10^{-2}\)), \texttt{Adam} with learning rate \(2.5{\times}10^{-5}\), exponential moving average, and no mixed precision; and PNDM \citep{liu2022pseudo} for sampling. For feature extraction we sample diffusion trajectories with PNDM using \(10\!-\!20\) steps (dataset-dependent), \(\eta{=}0\), and guidance scale \(1\) (unconditional).

\paragraph{DISC feature construction.}
For each input \(x\) we form a feature vector \(s(x)\) by aggregating metrics across timesteps \(t\in\mathcal{T}\). On CIFAR-10 we set \(|\mathcal{T}|=20\). On ImageNet we subsample \(\mathcal{T}\) to \(|\mathcal{T}|=10\) from a \(3000\)-step grid. At each \(t\) we compute: 
(i) \emph{reconstruction fidelity} between \(x\) and \(\hat x_t\): MSE, SSIM (Kornia defaults \citep{riba2020kornia}; \(11{\times}11\) window) and LPIPS (AlexNet backbone); 
(ii) \emph{score/trajectory consistency}, using  local-complexity (LC) construction around noised images at timestep \(t\), hyperplane/intersection counts following the original paper.

\paragraph{Baseline OOD detectors and settings.}
We use the reference implementations provided by the respective packages (primarily \texttt{pytorch-ood}, and official author repositories where applicable) and run each detector with the \emph{default} hyperparameters used in state-of-the-art benchmark suites. When a method requires calibration (e.g., ODIN temperature/perturbation), we follow the package’s standard ID-only validation protocol and reuse the resulting setting across all OOD families. No additional per-family tuning or deviations from the released implementations were introduced.

\paragraph{Evaluation protocols.}
For binary ID–vs–OOD detection we report AUROC. Any fixed thresholds \(\tau\) are chosen on ID validation. For optional multi-OOD classification, we (i) run unsupervised K-means with \(k\) equal to the number of OOD families on either scalar scores or DISC vectors and perform Hungarian alignment for cluster-to-family mapping; and (ii) train a small supervised MLP (2 hidden layers, width \(256\), ReLU, dropout \(0.2\)). Optional pairwise OOD–vs–OOD separability is computed as AUROC on the chosen features. Unless otherwise noted, we report mean\(\pm\)std over \(N=5\) seeds; where confidence intervals are shown we use nonparametric bootstrap over samples.

\paragraph{Computing environment and reproducibility.}
All experiments run on a single NVIDIA GPU (\(\ge\)16\,GB VRAM). We use \texttt{PyTorch} (2.1), CUDA (12.2), and most notably the following libraries: \texttt{pytorch-ood} \citep{kirchheim2022pytorch},  \texttt{Kornia}  \citep{riba2020kornia}, \texttt{PyWavelets} \citep{Lee2019}, \texttt{MONAI Generative} \citep{pinaya2023generative}, and \texttt{improved-diffusion} \citep{nichol2021improved}. 

\paragraph{Hyperparameters (summary).}
Full hyperparameters are recorded in the configuration files. In brief:  
\emph{CIFAR-10 (images).} Diffusion scheduler with \(T{=}1000\) training steps (MONAI DDPM), linear \(\beta\)-schedule (\(10^{-4}\!\to\!2{\times}10^{-2}\)), PNDM sampler with \(|\mathcal{T}|=20\) evaluation timesteps; base lr \(2.5{\times}10^{-5}\); batch size \(128\).  
\emph{ImageNet-1k (images).} Improved-diffusion U-Net with \(T{=}4000\) and cosine schedule; LC/evaluation uses \(|\mathcal{T}|=10\) subsampled timesteps; batch size \(128\).  
\emph{DISC metrics.} SSIM uses Kornia defaults (window \(11{\times}11\)), LPIPS backbone \texttt{alex}, score-step \(\Delta=1\), LBP hist \(256\) bins, edge-orientation \(36\) bins, DCT/wavelet hist \(50\) bins, wavelet \texttt{haar} (level 1).  
\emph{Baselines.} ODIN \(\epsilon=0.002\) (temperature tuned on ID validation), Mahalanobis layer = penultimate avg-pool (no shrinkage unless specified), DICE \(p=0.65\), ViM \(d=64\), MCD \(30\) samples. A single tuned setting per detector is reused across OOD families.

\smallskip
All remaining implementation details are included in the code release to ensure exact reproducibility.
\section{Additional Results}
\label{sec:add_results}

This appendix augments the main text with four sets of results. 
First, we provide additional score distributions and two–dimensional embeddings (t-SNE) on \textbf{CIFAR-10} and \textbf{ImageNet} to visualize how ID and OOD families separate under different detectors. 
Second, we report ImageNet-wide AUROC tables that cover both standard baselines and diffusion-derived features, enabling a fine-grained assessment of \textbf{DISC} across OOD types. 
Third, we ablate \textbf{DISC} by varying the underlying metric operators (reconstruction fidelity, texture/geometry, and path-consistency) and quantify their individual and combined contributions. 
Finally, we present comprehensive \textbf{tabular} benchmarks with average AUROC and rank, per-dataset breakdowns, and qualitative embeddings. 
 We keep the same experimental protocol as  the main body.

\vspace{0.5em}
\subsection{Image Data: Extended Baselines}
\label{sec:add_results:cifar}

\begin{figure*}[htbp]
  \centering
  \begin{subfigure}[t]{0.32\textwidth}\centering
    \includegraphics[width=\linewidth]{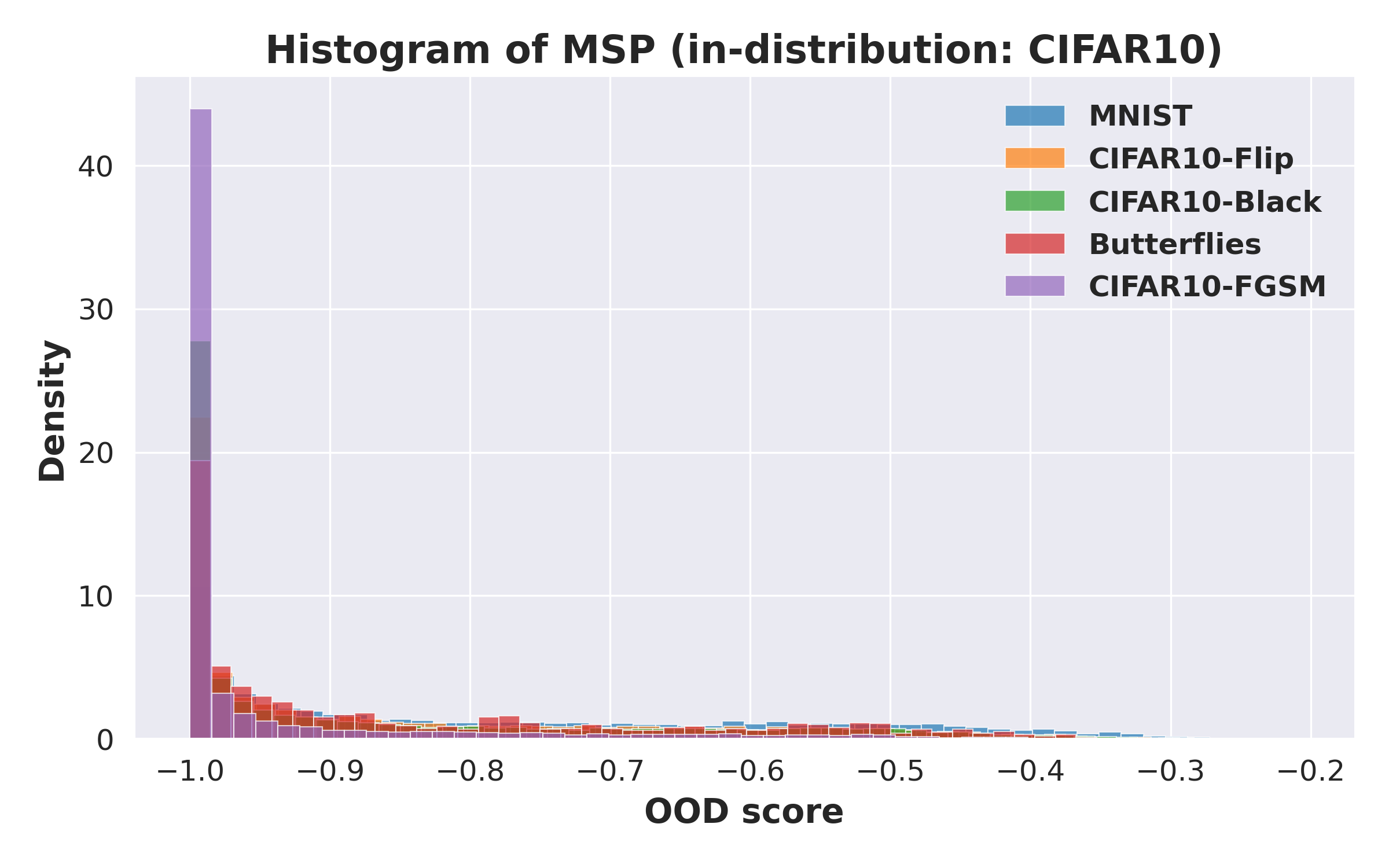}\caption{MSP}\end{subfigure}\hfill
  \begin{subfigure}[t]{0.32\textwidth}\centering
    \includegraphics[width=\linewidth]{AISTATS2026PaperPack/Figures/section3/baselines/CIFAR10_histograms/EnergyBased.png}\caption{Energy}\end{subfigure}\hfill
  \begin{subfigure}[t]{0.32\textwidth}\centering
    \includegraphics[width=\linewidth]{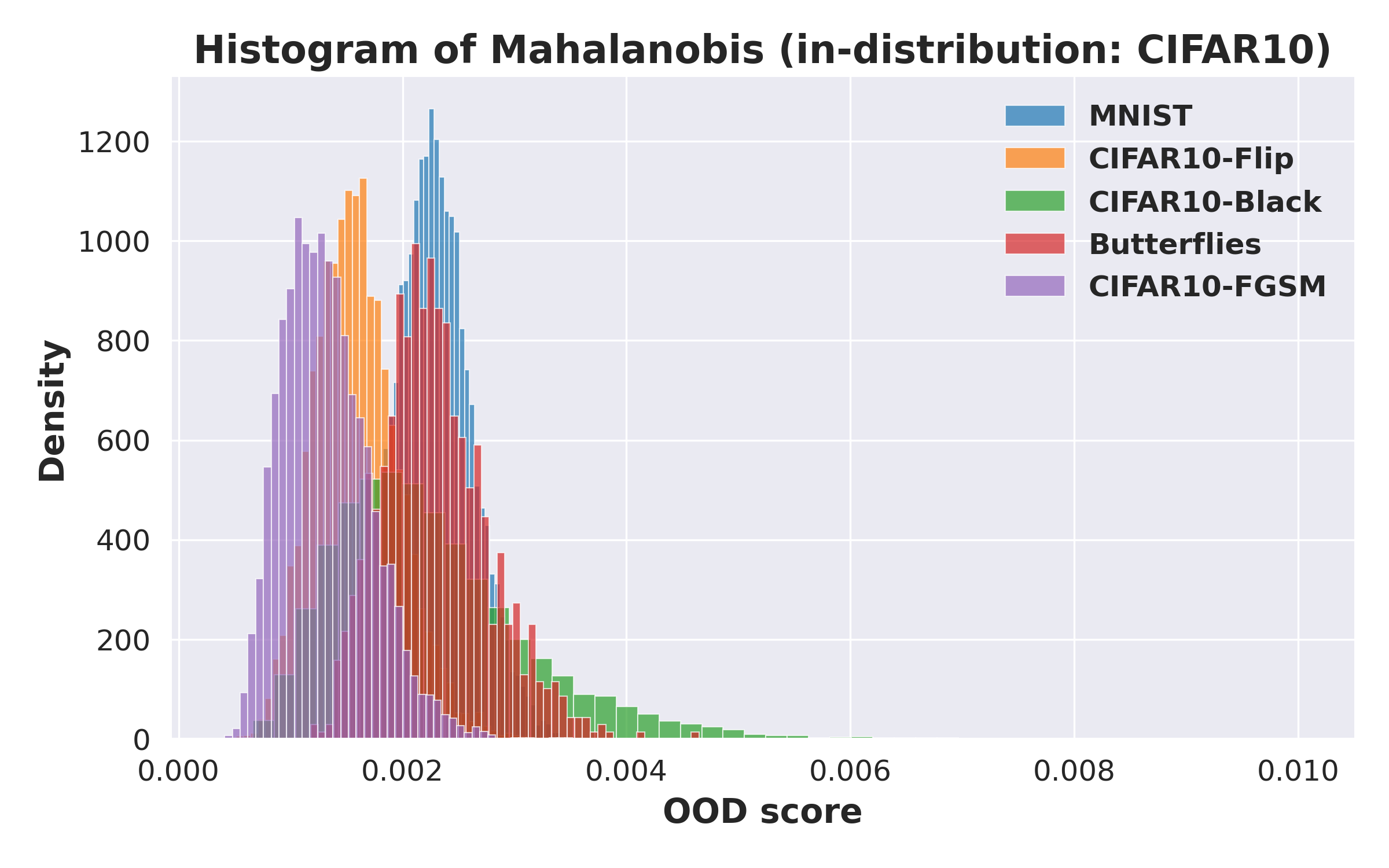}\caption{Mahalanobis}\end{subfigure}

  \vspace{0.25em}
  \begin{subfigure}[t]{0.32\textwidth}\centering
    \includegraphics[width=\linewidth]{AISTATS2026PaperPack/Figures/section3/baselines/CIFAR10_histograms/Mahalanobis+ODIN.png}\caption{Maha+ODIN}\end{subfigure}\hfill
  \begin{subfigure}[t]{0.32\textwidth}\centering
    \includegraphics[width=\linewidth]{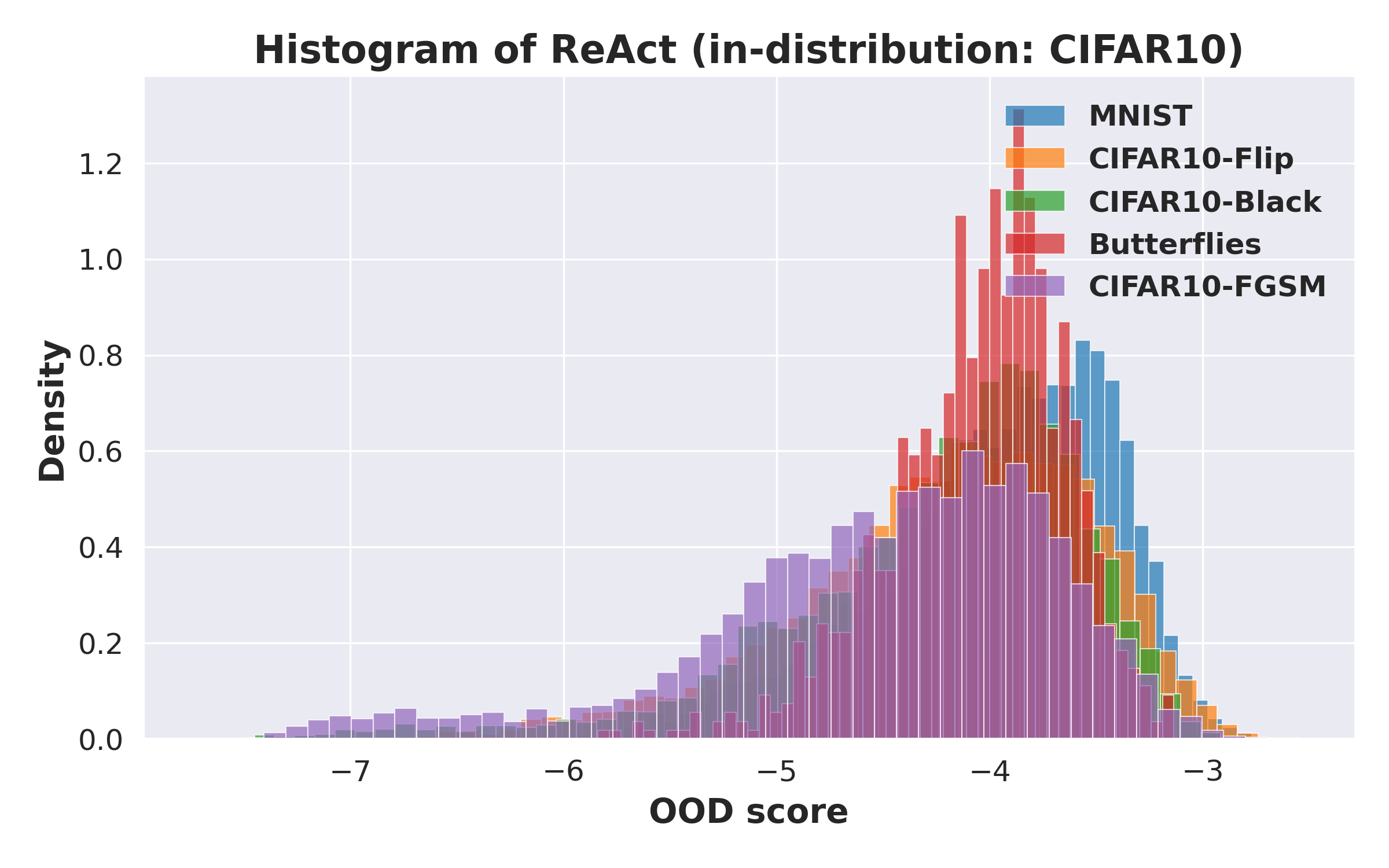}\caption{ReAct}\end{subfigure}\hfill
  \begin{subfigure}[t]{0.32\textwidth}\centering
    \includegraphics[width=\linewidth]{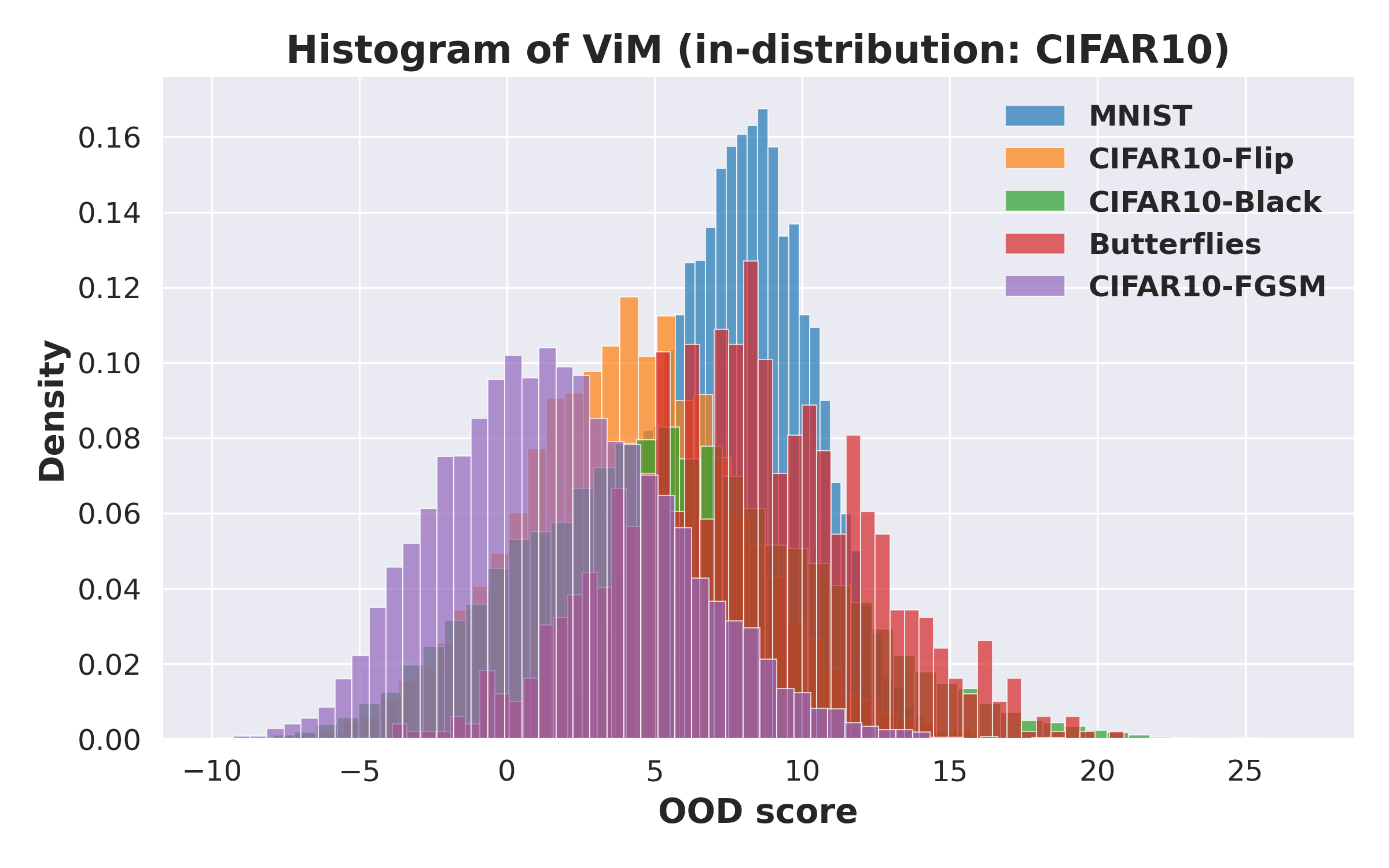}\caption{ViM}\end{subfigure}

  \vspace{0.25em}
  \begin{subfigure}[t]{0.32\textwidth}\centering
    \includegraphics[width=\linewidth]{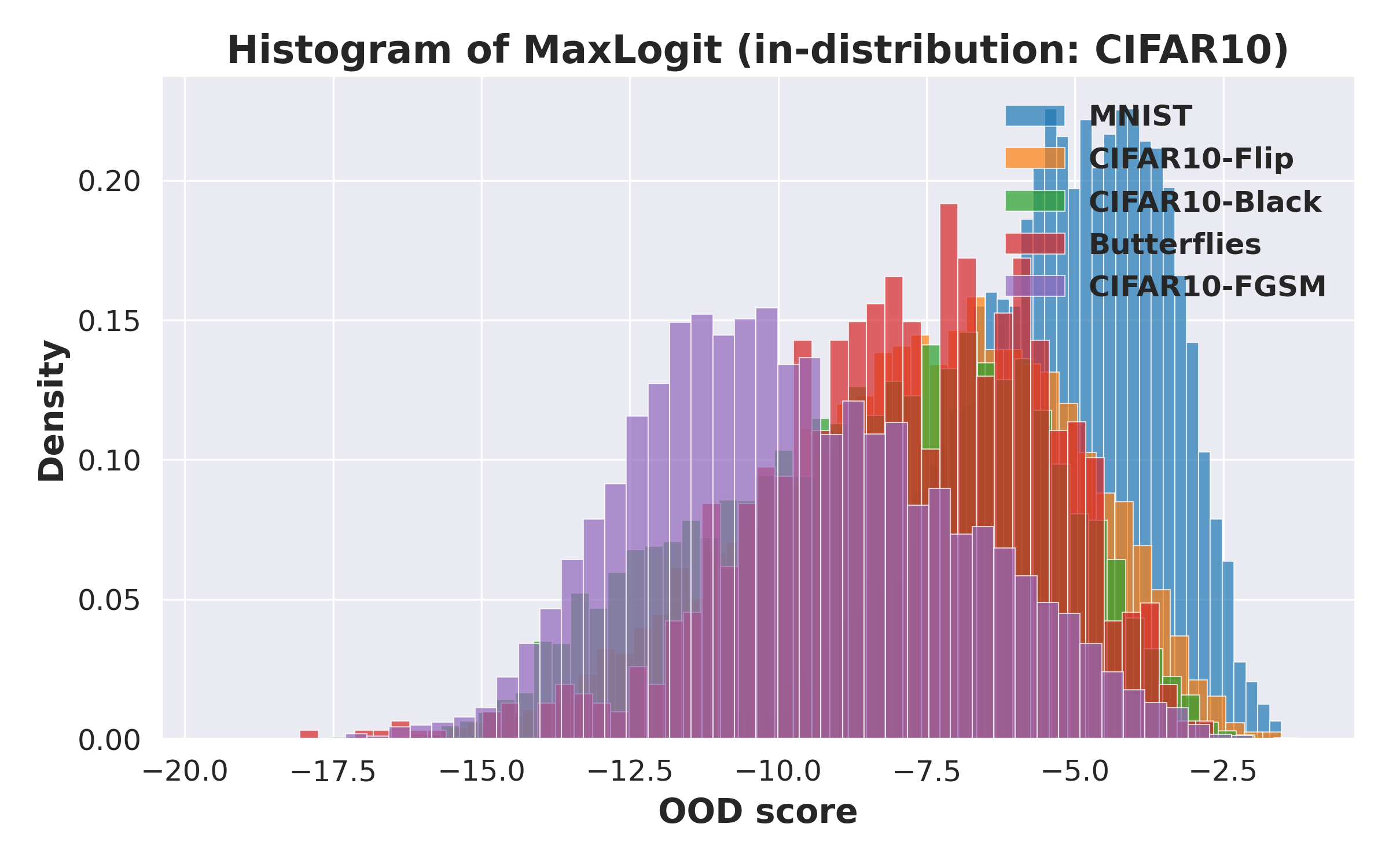}\caption{MaxLogit}\end{subfigure}\hfill
  \begin{subfigure}[t]{0.32\textwidth}\centering
    \includegraphics[width=\linewidth]{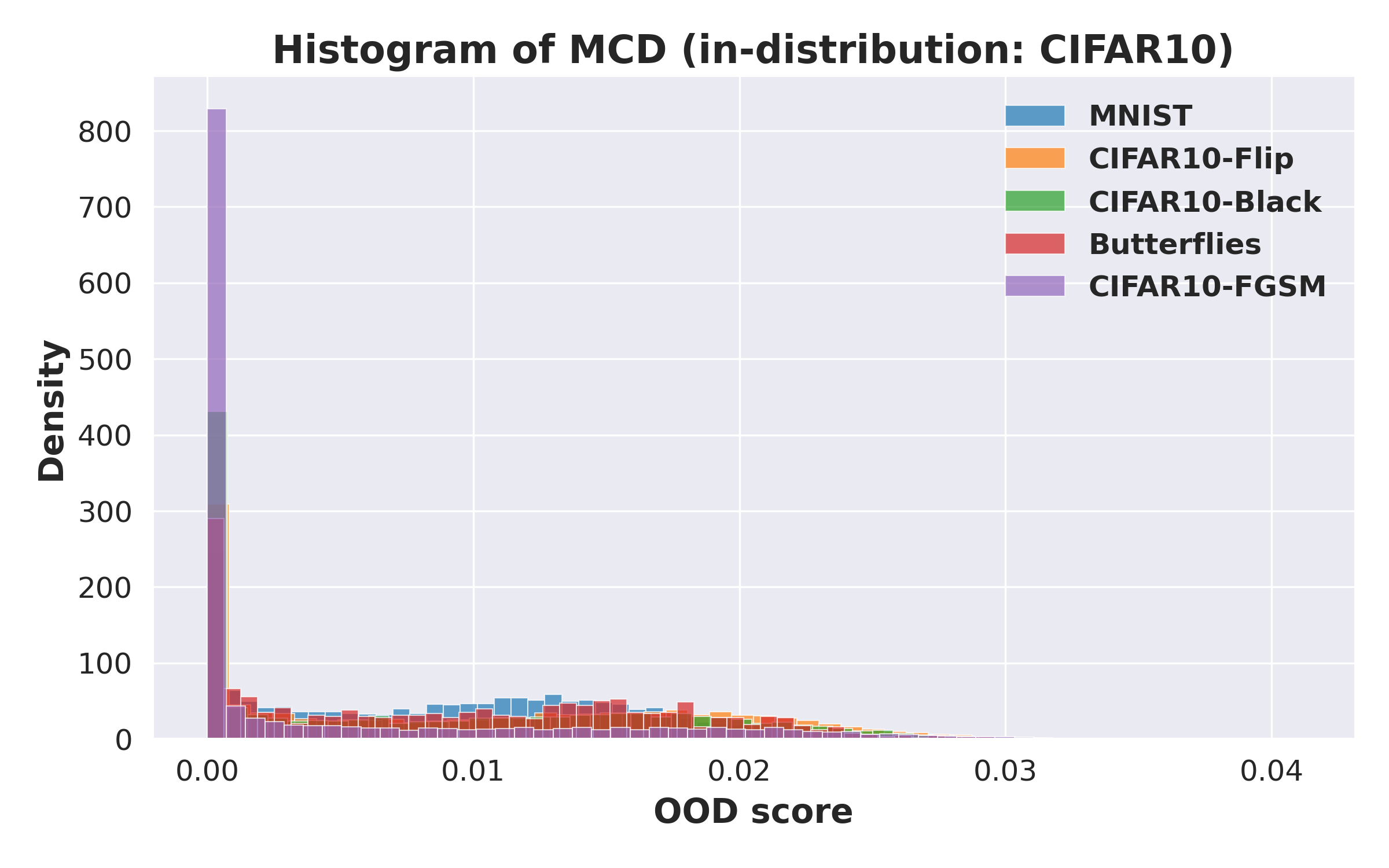}\caption{MC Dropout}\end{subfigure}\hfill
  \begin{subfigure}[t]{0.32\textwidth}\centering
    \includegraphics[width=\linewidth]{AISTATS2026PaperPack/Figures/section3/baselines/CIFAR10_histograms/SHE.png}\caption{SHE}\end{subfigure}

  \caption{Extended OOD score distributions on \textbf{CIFAR-10} for nine baseline detectors. Near-OOD families shift modestly from the ID mode; semantic OODs shift further yet overlap with one another and with some near-OODs. Densities are normalized and share identical binning.}
  \label{fig:cifar:score_distributions_ext}
\end{figure*}

\paragraph{More score histograms.}
Figure~\ref{fig:cifar:score_distributions_ext} extends the histogram analysis in the main body to nine representative detectors on CIFAR-10. Across all methods, OOD families shift away from the ID peak but \emph{retain substantial overlap}. Near-distribution shifts (\texttt{CIFAR-FGSM}, \texttt{CIFAR-Flip}, central black occlusion) stay closest to the ID mode, while semantic OODs (\texttt{Butterflies}, \texttt{MNIST}) move further into the high-novelty tail. The relative ordering of families is consistent across detectors, replicating the main-text observation that a single scalar score is often sufficient for ID-vs-OOD but not for \emph{family-aware} separation.

\vspace{0.5em}
\label{sec:add_results:imagenet}

\begin{figure*}[htbp]
  \centering
  \begin{subfigure}[t]{0.32\textwidth}\centering
    \includegraphics[width=\linewidth]{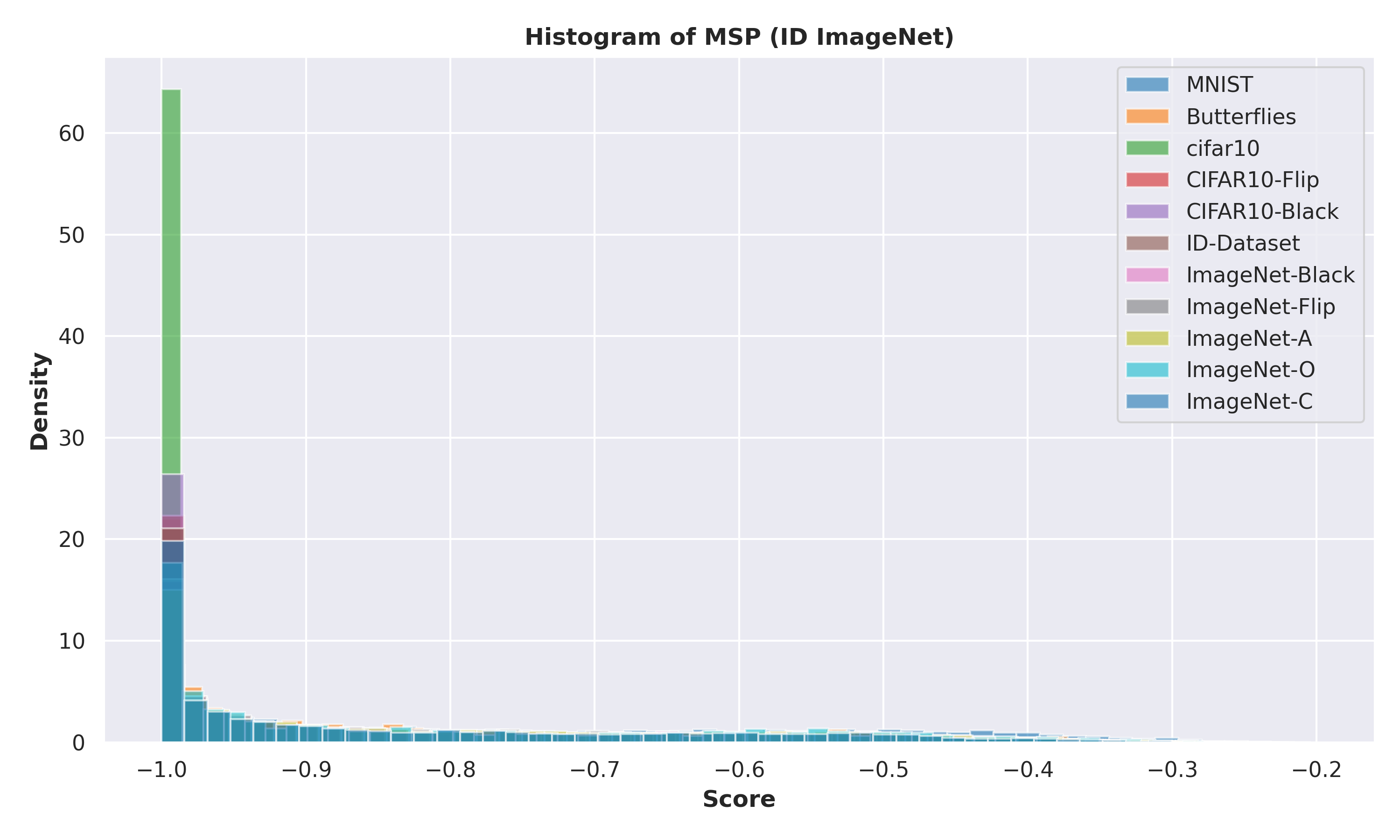}\caption{MSP}\end{subfigure}\hfill
  \begin{subfigure}[t]{0.32\textwidth}\centering
    \includegraphics[width=\linewidth]{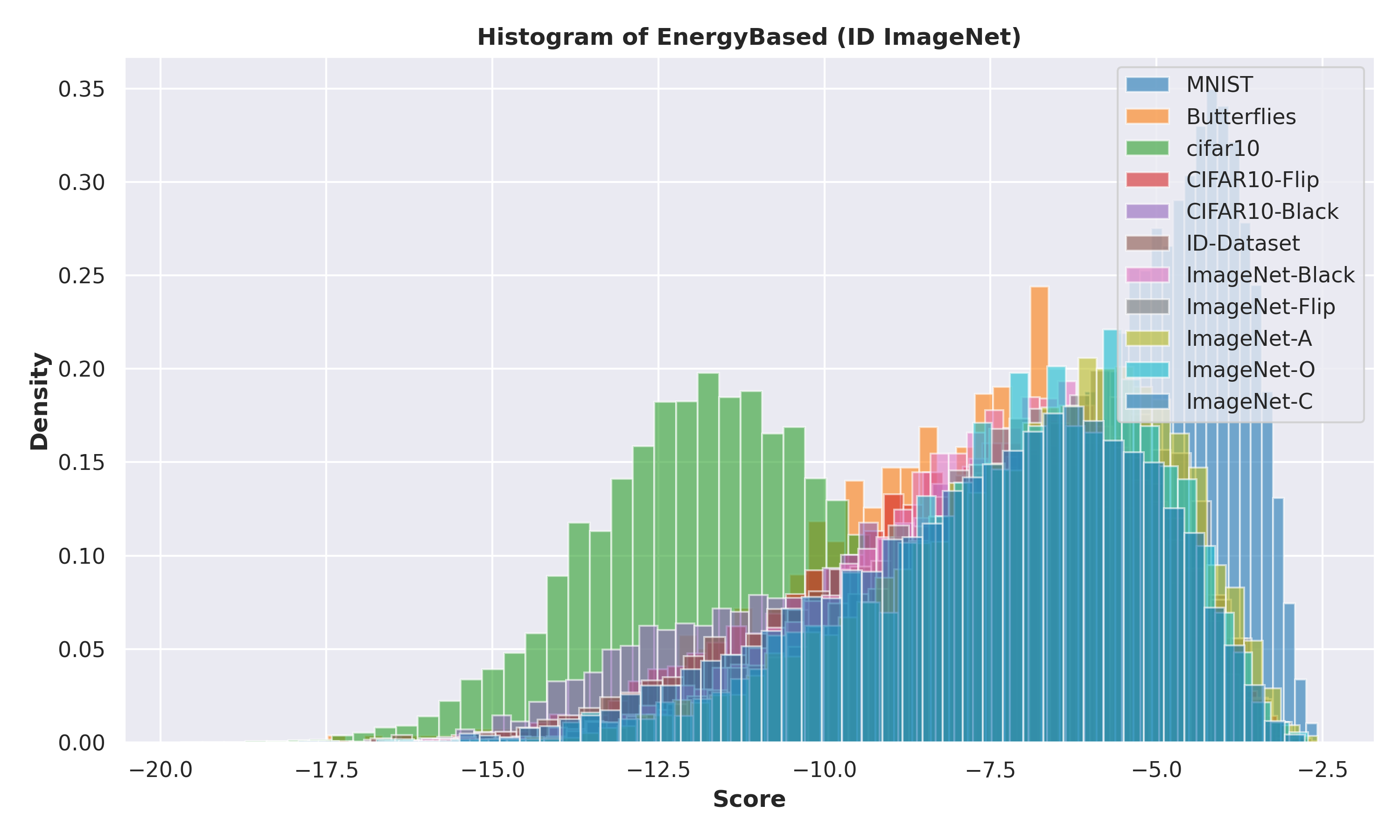}\caption{Energy}\end{subfigure}\hfill
  \begin{subfigure}[t]{0.32\textwidth}\centering
    \includegraphics[width=\linewidth]{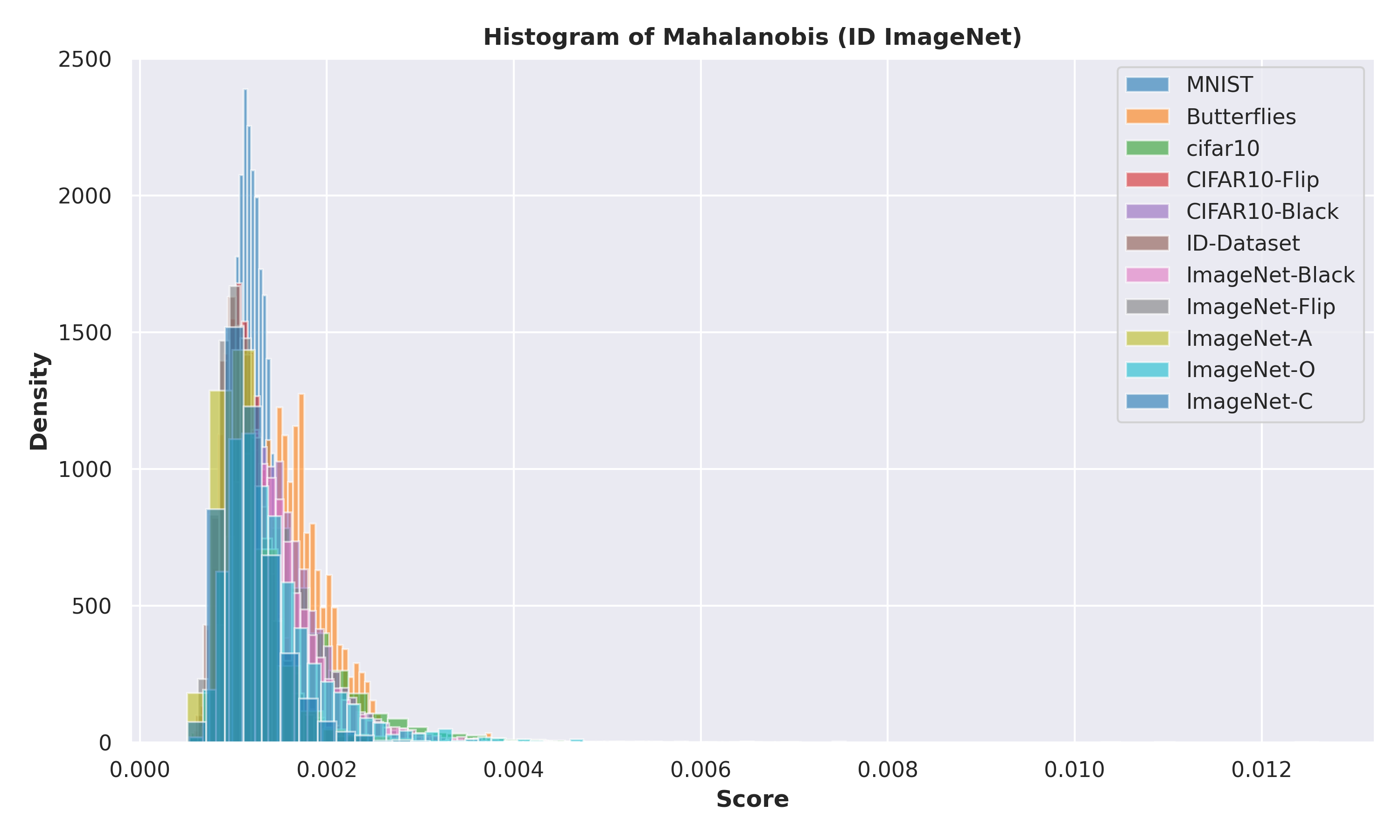}\caption{Mahalanobis}\end{subfigure}

  \vspace{0.25em}
  \begin{subfigure}[t]{0.32\textwidth}\centering
    \includegraphics[width=\linewidth]{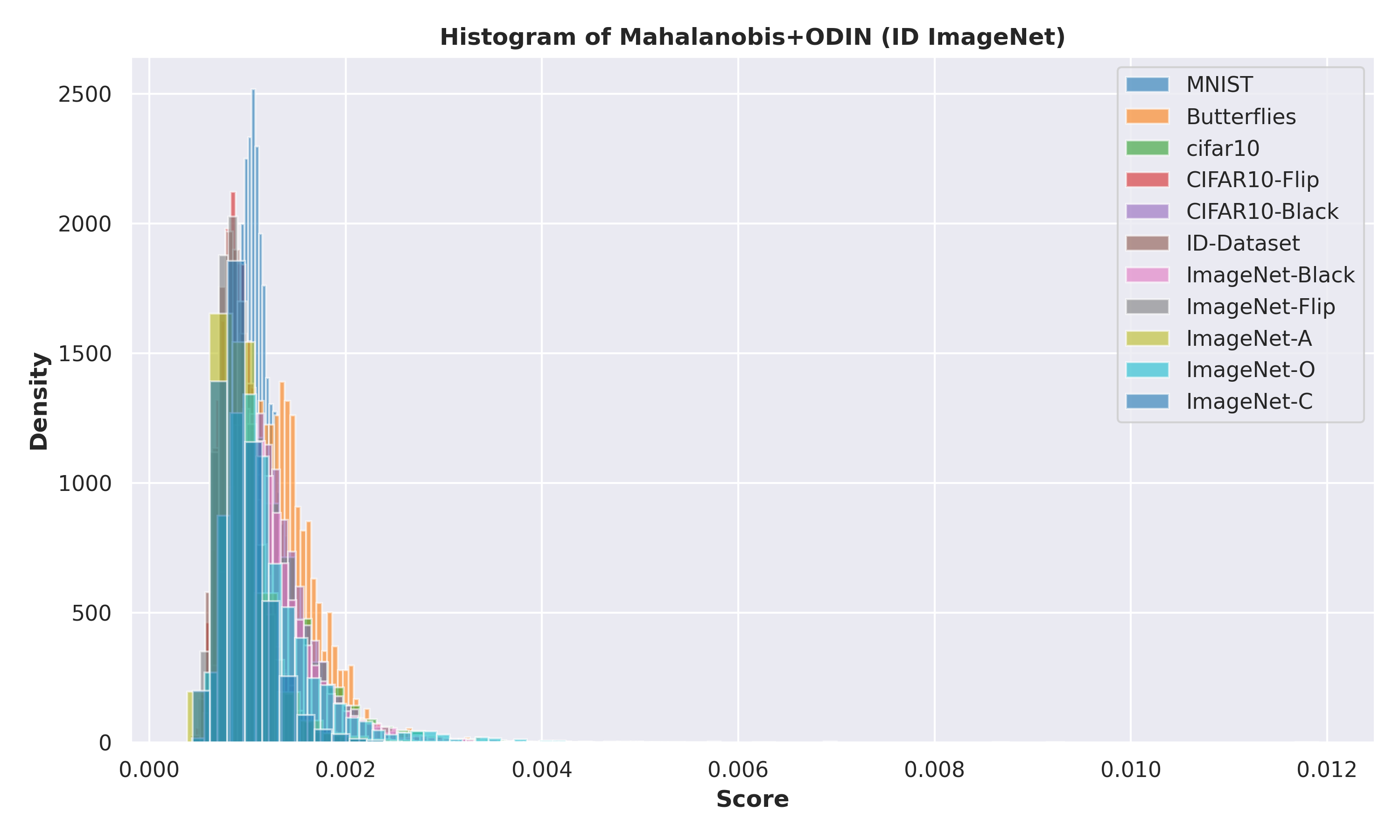}\caption{Maha+ODIN}\end{subfigure}\hfill
  \begin{subfigure}[t]{0.32\textwidth}\centering
    \includegraphics[width=\linewidth]{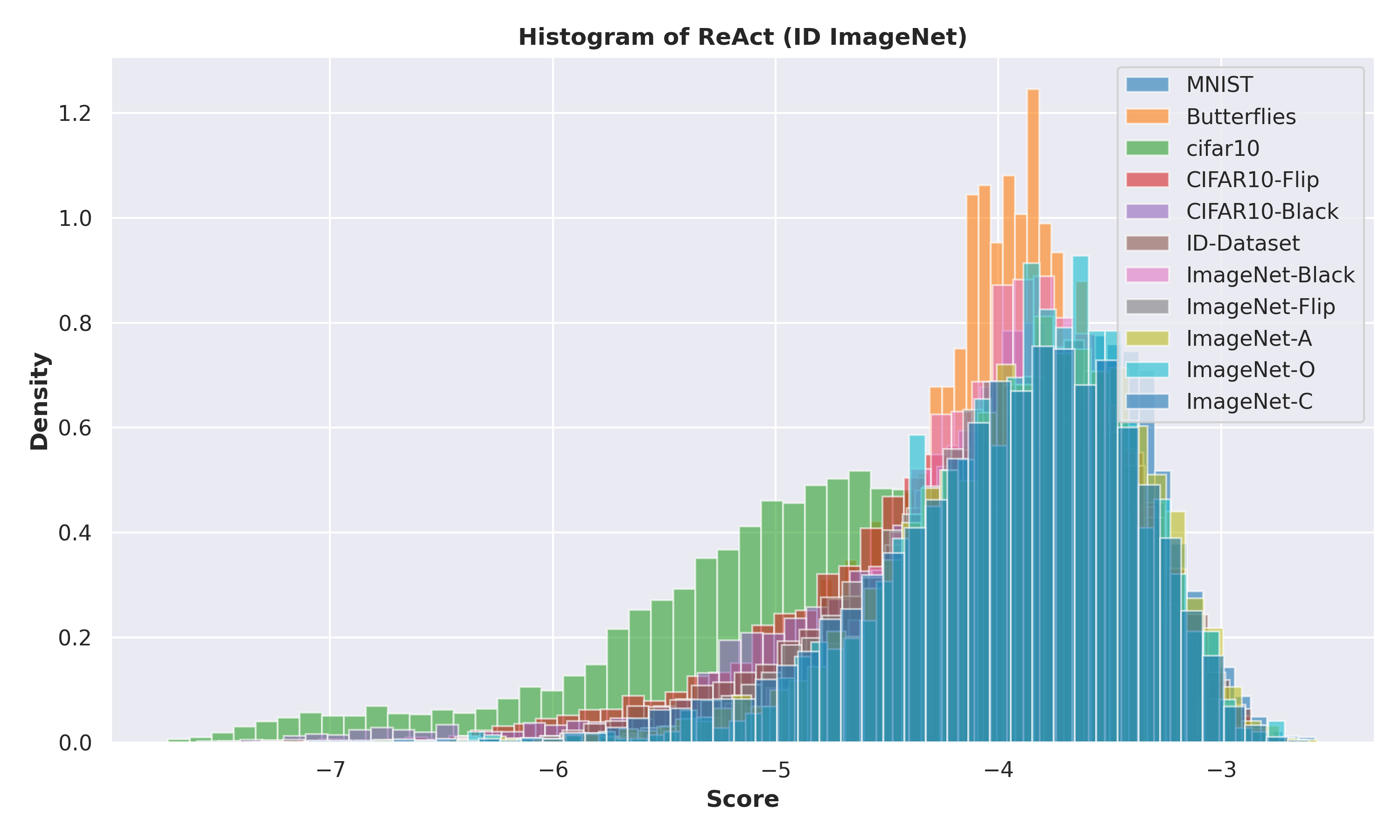}\caption{ReAct}\end{subfigure}\hfill
  \begin{subfigure}[t]{0.32\textwidth}\centering
    \includegraphics[width=\linewidth]{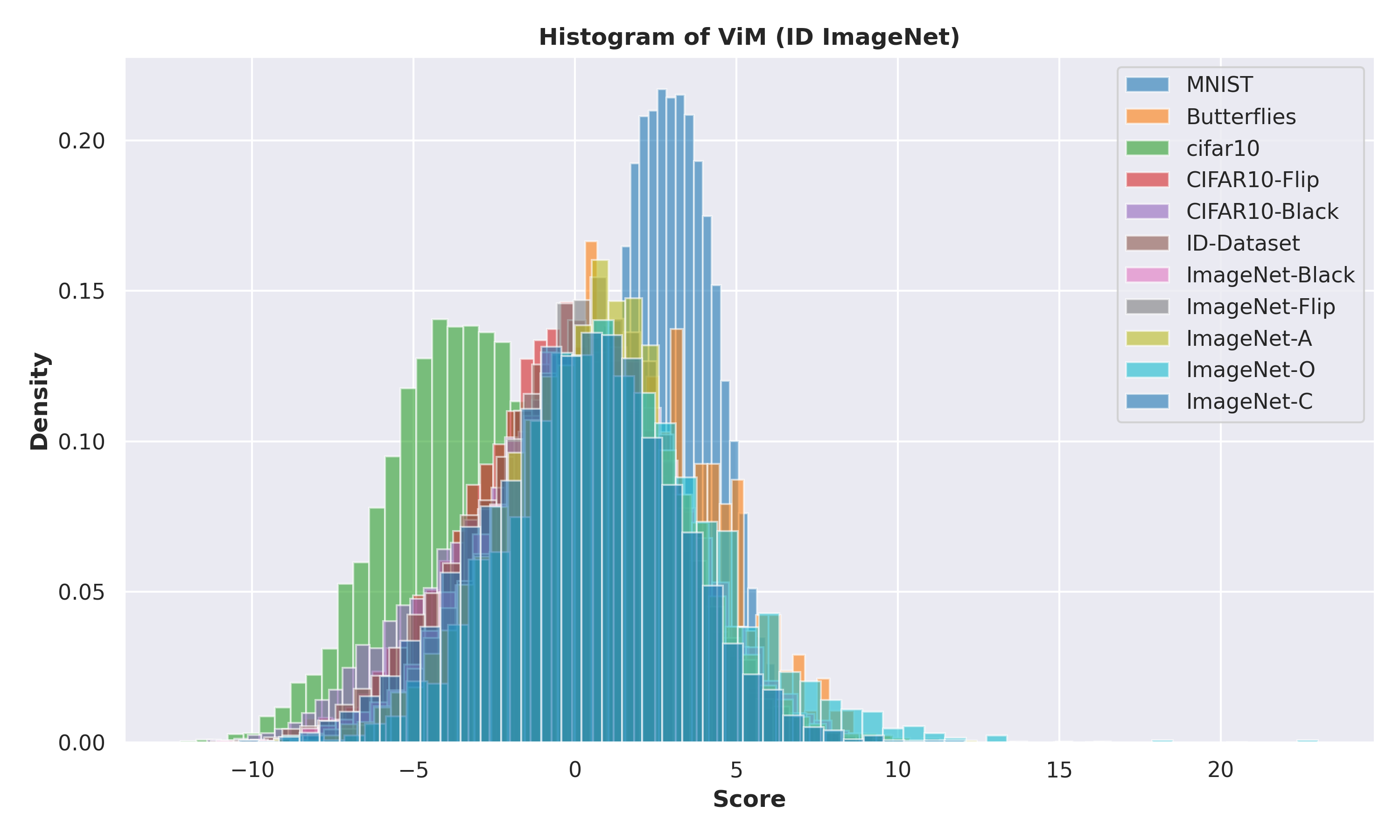}\caption{ViM}\end{subfigure}

  \vspace{0.25em}
  \begin{subfigure}[t]{0.32\textwidth}\centering
    \includegraphics[width=\linewidth]{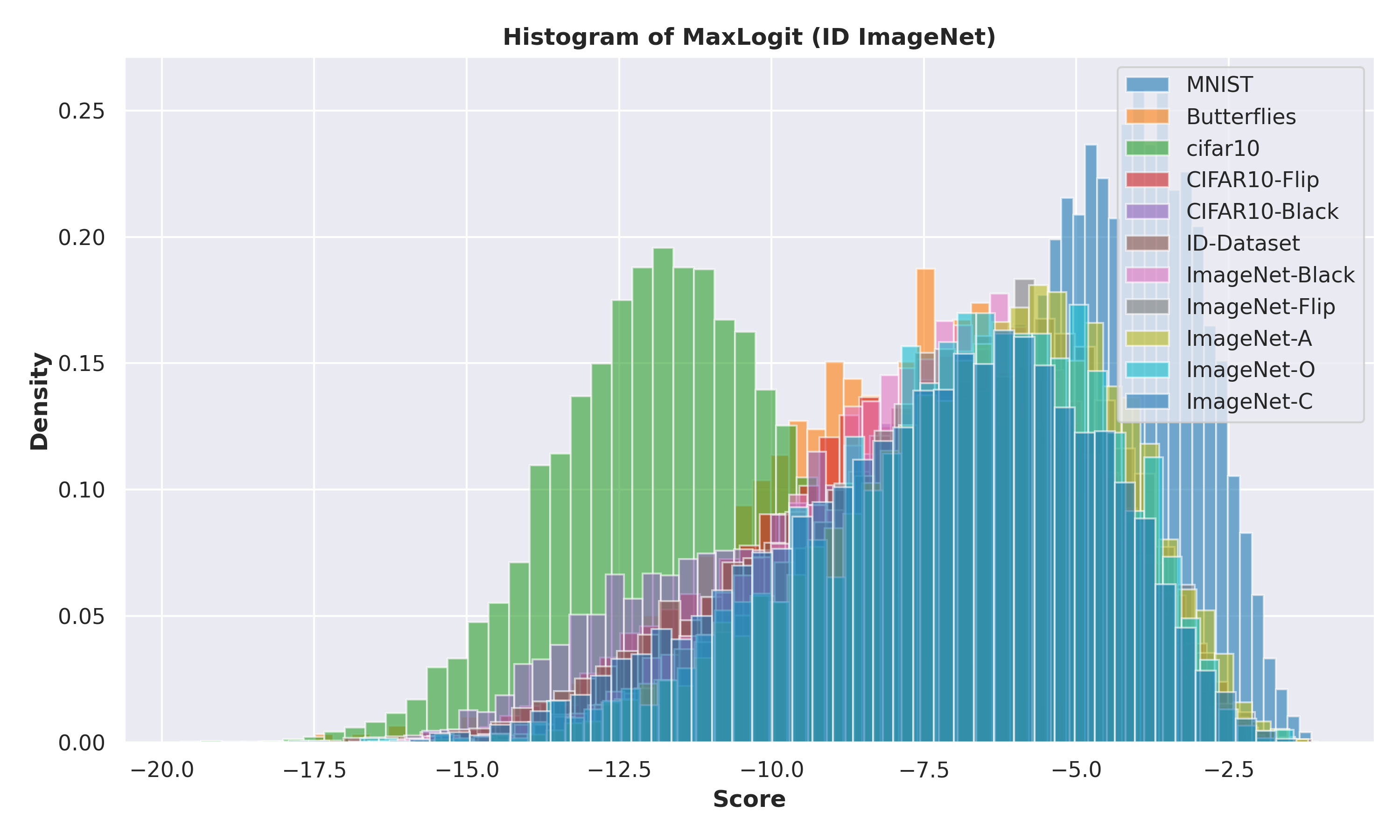}\caption{MaxLogit}\end{subfigure}\hfill
  \begin{subfigure}[t]{0.32\textwidth}\centering
    \includegraphics[width=\linewidth]{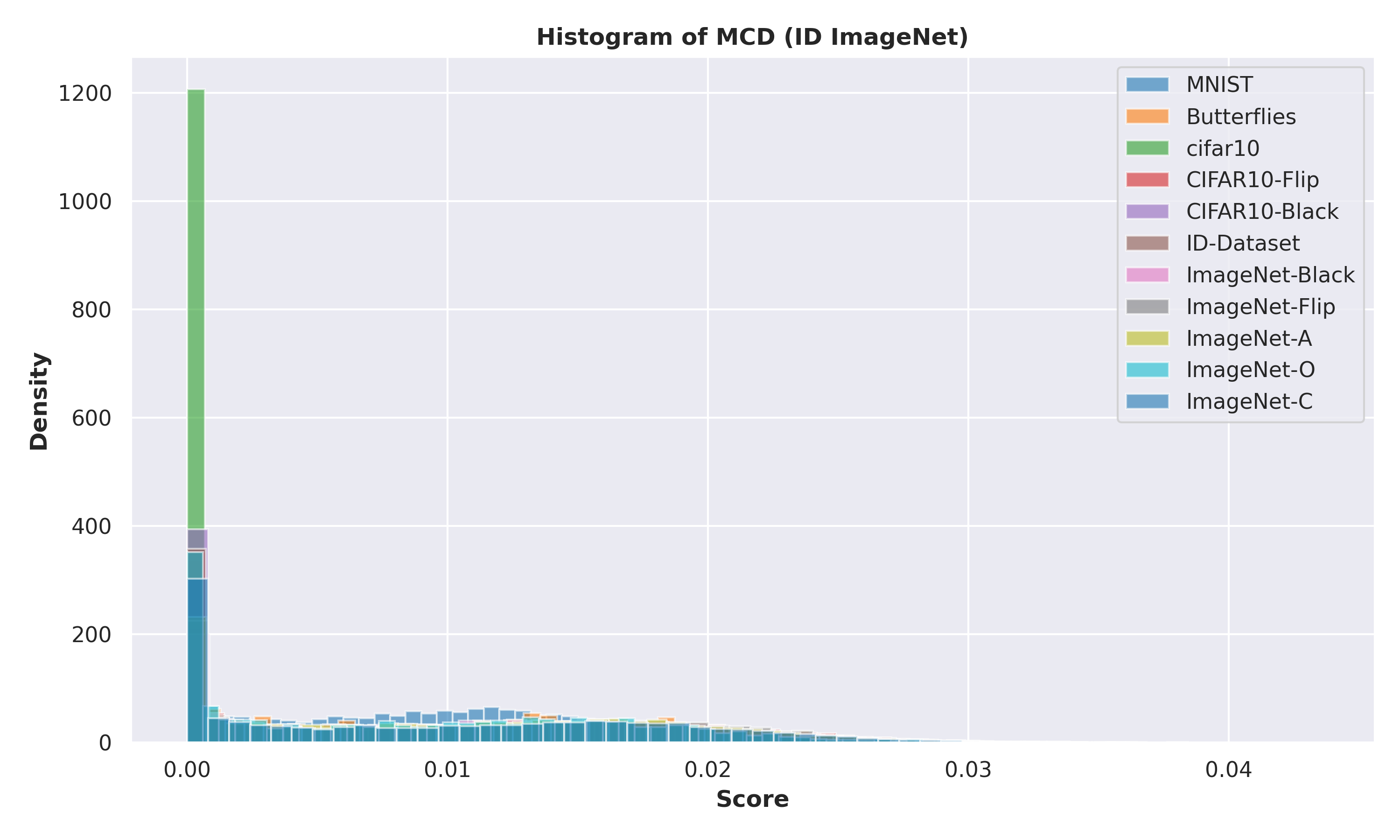}\caption{MC Dropout}\end{subfigure}\hfill
  \begin{subfigure}[t]{0.32\textwidth}\centering
    \includegraphics[width=\linewidth]{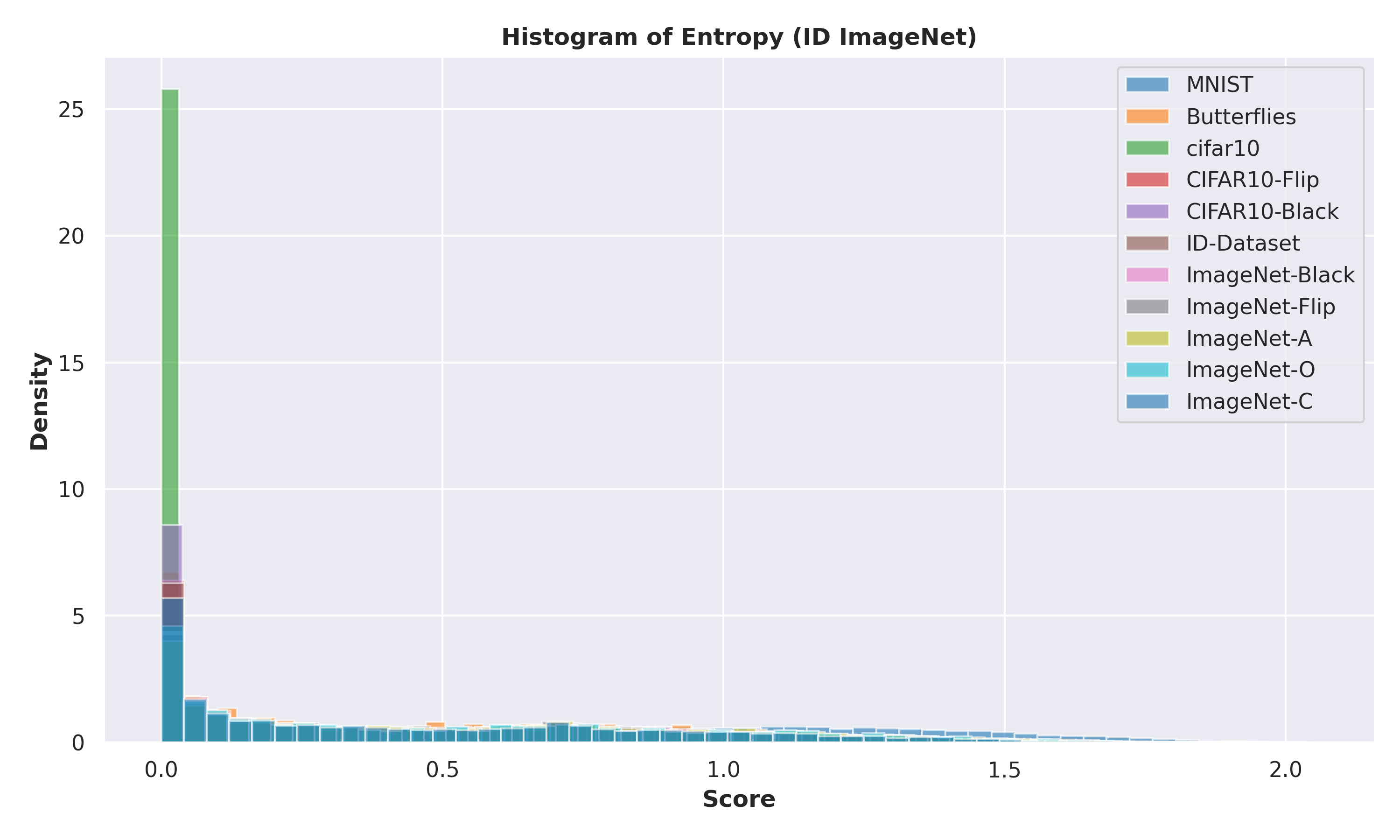}\caption{Entropy}\end{subfigure}

  \caption{Extended OOD score distributions on \textbf{ImageNet}. Family ordering is broadly consistent across detectors; the overlap among natural-image OODs remains pronounced.}
  \label{fig:imageNet:score_distributions_ext}
\end{figure*}

\paragraph{Score histograms.}
Figure~\ref{fig:imageNet:score_distributions_ext} expands the score–distribution view on \textbf{ImageNet}. Across detectors the same pattern holds: every OOD family shifts away from the ID mode, but natural-image OODs (\texttt{ImageNet-A/O/C}) occupy essentially the \emph{same} score range as ID, producing heavy overlap and long shared tails. Detectors that operate directly on prediction confidence (MSP/Entropy/MaxLogit) display highly skewed, low-variance ID peaks with OOD mass bleeding into the shoulder; Energy behaves similarly but with a broader dynamic range, which makes far-domain OODs (\texttt{MNIST}, \texttt{Butterflies}) accumulate in the high-novelty tail. Feature–space methods (Mahalanobis and Maha+ODIN) sharpen the ID spike and pull the near-distribution transforms (flips/occlusions) closer to it, while still separating the extreme-domain OODs; ODIN’s temperature/perturbation mainly reduces overlap in the immediate ID shoulder. ViM and ReAct alter calibration rather than class order: they shift all curves but preserve the relative difficulty of families. 

\begin{figure*}[htbp]
  \centering
  \begin{subfigure}[t]{0.32\textwidth}\centering
    \includegraphics[width=\linewidth]{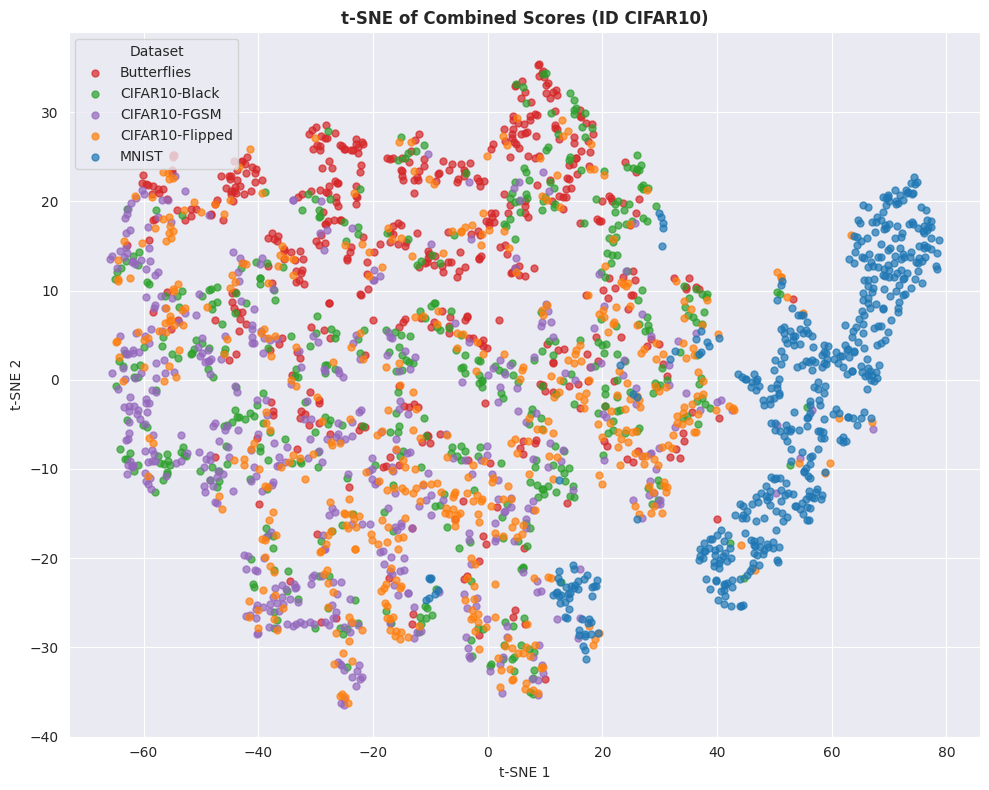}\caption{CIFAR}\end{subfigure}
  \begin{subfigure}[t]{0.32\textwidth}\centering
    \includegraphics[width=\linewidth]{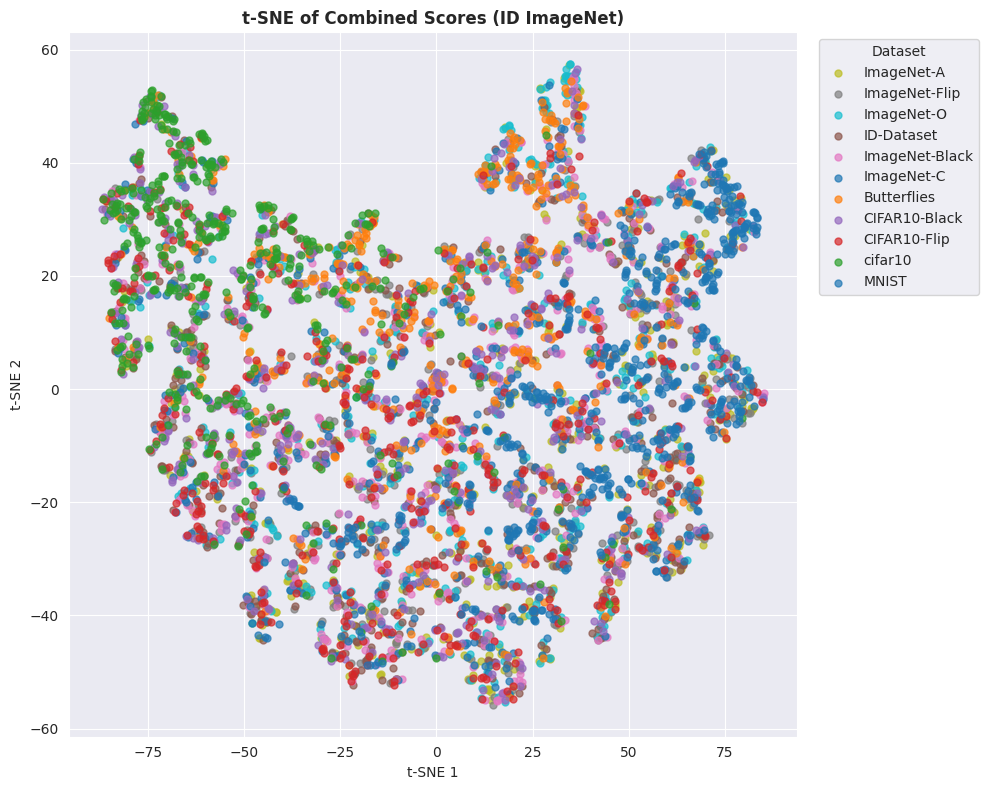}\caption{ImageNet}\end{subfigure}\hfill
  
  \caption{t-SNE of detector features/scores on \textbf{ImageNet}. Far OODs form more distinct clusters than on CIFAR-10, but natural corruptions and near-distribution variants still overlap across detectors. t-SNE of detector features/scores on \textbf{CIFAR-10}. Far OODs separate clearly, but near-OOD families entangle with the ID manifold across most detectors, underscoring the limits of single-score detectors for family-aware analysis}
  \label{fig:imagenet:tsne_all}
\end{figure*}

\paragraph{t-SNE embeddings across detectors.}
Figure~\ref{fig:imagenet:tsne_all} visualizes the joint embedding of \emph{combined detector scores} for CIFAR-10 (a) and ImageNet (b). On \textbf{CIFAR-10}, the far-shift \texttt{MNIST} forms a compact island well separated from the ID cloud, while \texttt{Butterflies} occupies a broad swath that is partly distinct yet still interleaves with ID at its boundary. In contrast, near-distribution transformations (\texttt{FGSM}, flips, central black-out) dissolve into the ID manifold and trace smooth arcs along its periphery, consistent with their small score deltas in the histograms. On \textbf{ImageNet}, the strongest separation again occurs for extreme domain shifts (\texttt{MNIST}, and to a lesser extent \texttt{CIFAR-10}), whereas natural-image families (\texttt{ImageNet-A/O/C}) largely overlap with the ID lobe and with each other, reflecting the difficulty of disentangling semantically similar content using a single signal. Taken together, the embeddings reinforce our central claim: even if scalar detectors are adequate for binary ID–vs–OOD they compress family structure, which makes fine grained detection very hard.

\begin{table}[ht]
\centering
\scriptsize
\setlength{\tabcolsep}{3pt}
\renewcommand{\arraystretch}{0.9}
\caption{AUROC results across OOD datasets for various methods with ImageNet as in-distribution dataset. Boldface in text discusses notable family-wise trends.}
\label{tab:combined_aurocs}
\resizebox{\textwidth}{!}{

\begin{tabular}{@{}lrrrrrrrrr@{}}
\toprule
Method & ImageNet-A & ImageNet-F & ImageNet-O & ImageNet-B & ImageNet-C & Butterflies & CIFAR10 & MNIST & AVG \\
\midrule

\multicolumn{10}{l}{\textbf{DDPM}:} \\
ALL (DISC)       & 0.607 & 0.507 & 0.658 & 0.767 & 0.557 & 0.939 & 0.801 & 0.989 & 0.728 \\
LBP       & 0.547 & 0.494 & 0.601 & 0.898 & 0.555 & 0.638 & 0.956 & 0.995 & 0.710 \\
LC        & 0.598 & 0.505 & 0.587 & 0.749 & 0.539 & 0.858 & 0.507 & 0.990 & 0.667 \\
MSE       & 0.544 & 0.505 & 0.611 & 0.482 & 0.530 & 0.983 & 0.558 & 0.989 & 0.650 \\
LPIPS     & 0.626 & 0.506 & 0.627 & 0.685 & 0.606 & 0.705 & 0.837 & 0.474 & 0.633 \\
SSIM      & 0.577 & 0.504 & 0.588 & 0.849 & 0.559 & 0.826 & 0.498 & 0.650 & 0.631 \\
HOG       & 0.609 & 0.495 & 0.558 & 0.490 & 0.469 & 0.377 & 0.527 & 0.748 & 0.534 \\
\midrule
\multicolumn{10}{l}{\textbf{OOD detectors}} \\
EnergyBased      & 0.592 & 0.564 & 0.561 & 0.510 & 0.516 & 0.436 & 0.154 & 0.826 & 0.506  \\
Entropy          & 0.575 & 0.556 & 0.552 & 0.566 & 0.516 & 0.501 & 0.147 & 0.694 & 0.506 \\
MCD              & 0.549 & 0.556 & 0.503 & 0.551 & 0.514 & 0.512 & 0.155 & 0.489 & 0.481  \\
MSP              & 0.570 & 0.554 & 0.552 & 0.560 & 0.516 & 0.508 & 0.150 & 0.663 & 0.502 \\
Mahalanobis      & 0.475 & 0.482 & 0.671 & 0.718 & 0.512 & 0.867 & 0.751 & 0.643 & 0.640 \\
Mahalanobis+ODIN & 0.472 & 0.480 & 0.690 & 0.721 & 0.503 & 0.880 & 0.743 & 0.763 & 0.653 \\
MaxLogit         & 0.591 & 0.563 & 0.561 & 0.516 & 0.516 & 0.443 & 0.154 & 0.810 & 0.506 \\
ODIN             & 0.540 & 0.537 & 0.612 & 0.610 & 0.482 & 0.524 & 0.184 & 0.873 & 0.533 \\
ReAct            & 0.556 & 0.515 & 0.549 & 0.516 & 0.523 & 0.468 & 0.219 & 0.560 & 0.478 \\
ViM              & 0.574 & 0.553 & 0.623 & 0.550 & 0.529 & 0.638 & 0.256 & 0.773 & 0.545 \\
\bottomrule
\end{tabular}
}
\end{table}

\paragraph{Per-dataset AUROC on ImageNet.}
Table~\ref{tab:combined_aurocs} lists AUROC by OOD family. Diffusion-derived features (\textbf{DISC}) achieve the best \emph{average} (\(\mathbf{0.728}\)), surpassing the strongest scalar baseline, Mahalanobis+ODIN (\(0.653\)), by \(+0.075\). The largest gains appear on texture/shape mismatches and occlusions—\texttt{Butterflies} \(0.983\) (MSE), \texttt{MNIST} \(0.989\), \texttt{CIFAR-10} \(0.956\) (LBP), and \texttt{ImageNet-Black} \(0.898\). For natural outliers (\texttt{ImageNet-O}), Mahalanobis+ODIN remains competitive (\(0.690\) vs.\ \(0.658\) for DISC), and flips favor calibrated confidence methods (MSP/Energy \(\sim0.56\) vs.\ DDPM \(\sim0.51\)). Overall, diffusion trajectories add complementary texture/geometry cues, while near-invariant transforms are still best handled by calibrated logits.
\vspace{0.5em}

\section{Additional Ablation Studies}
\label{sec:appendix_ablations}

\paragraph{Ablation over metric groups.} To assess whether the full metric suite is both necessary and beneficial for DISC, we performed a leave-one-out ablation over the different metric groups. Results averaged across 5 random seeds are reported in Table~\ref{tab:metric_ablation}.

The results show that removing any individual metric group leads to a drop in supervised accuracy. The largest decreases are observed when excluding LPIPS or Wavelets, suggesting that these components contribute most strongly to the supervised setting. In the unsupervised setting, measured via clustering accuracy, performance declines are more evenly distributed across metric groups. The main exception is SSIM: removing SSIM yields a small increase of roughly 1\% for K-means clustering. Overall, these findings support the value of the proposed multi-statistic embedding and indicate that combining complementary metric groups leads to the most robust OOD characterization.

\begin{table}[h!]
\centering
\caption{Leave-one-out ablation of the DISC metric groups. We report clustering accuracy (Clust.\ Acc.) and supervised accuracy (Sup.\ Acc.), together with standard deviations over 5 runs and the relative change ($\Delta$) with respect to the full metric suite.}
\label{tab:metric_ablation}
\resizebox{\textwidth}{!}{
\begin{tabular}{lcccccc}
\toprule
\textbf{Combination} & \textbf{Clust. Acc} & \textbf{Clust. Acc Std} & \textbf{Sup. Acc} & \textbf{Sup. Acc Std} & \textbf{$\Delta$ Clust. Acc} & \textbf{$\Delta$ Sup. Acc} \\
\midrule
All Combined       & 0.4911 & 0.0118 & 0.6203 & 0.0163 & 0       & 0       \\
All except SSIM    & 0.5020 & 0.0148 & 0.5957 & 0.0137 & +0.0108 & -0.0246 \\
All except MSE     & 0.4623 & 0.0173 & 0.5950 & 0.0204 & -0.0289 & -0.0253 \\
All except LBP     & 0.4456 & 0.0196 & 0.5924 & 0.0103 & -0.0456 & -0.0279 \\
All except LPIPS   & 0.4794 & 0.0124 & 0.5911 & 0.0130 & -0.0117 & -0.0292 \\
All except Wavelets & 0.4694 & 0.0095 & 0.5870 & 0.0093 & -0.0217 & -0.0333 \\
All except LC      & 0.4894 & 0.0075 & 0.6170 & 0.0064 & -0.0017 & -0.0033 \\
\bottomrule
\end{tabular}
}
\end{table}

\label{sec:add_results:disc_imagenet}

Table~\ref{tab:perf_three_methods} compares metric groups under unsupervised (KMeans/GMM) and supervised (MLP) multi-OOD classification. Supervision consistently boosts performance, and \textbf{combining all metrics} yields the best accuracy (MLP \(0.625\)), confirming complementarity across reconstruction fidelity, local complexity, and texture operators.

\begin{figure}[ht]
    \centering
    \includegraphics[width=0.5\linewidth]{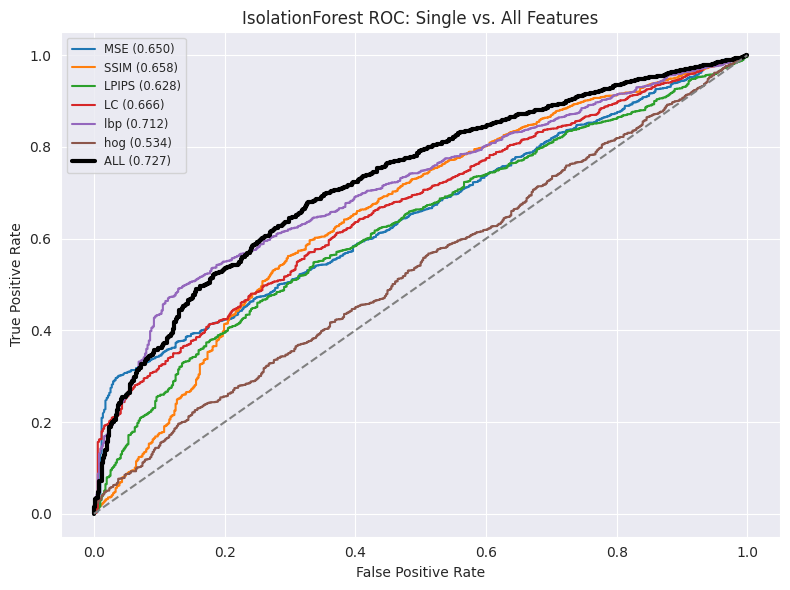}
    \caption{Sensitivity of an isolation-forest head to different DISC metric subsets on ImageNet (higher is better). Trends mirror Table~\ref{tab:perf_three_methods}: texture and LC are strong standalone; the full combination is best.}
    \label{fig:iforestonMetrics}
\end{figure}

\begin{table}[ht]
\centering
\caption{ Classification accuracy of OOD Types using different Test Statistics. Experiments repeated 5 times and average results are reported.}
\label{tab:perf_three_methods}

\begin{tabular}{@{}lccc@{}}
\toprule
\textbf{Method} & \textbf{KMeans} & \textbf{GMM} & \textbf{MLP} \\
\midrule
MSE    & 0.3275 & 0.3594 & 0.3905 \\
SSIM   & 0.3728 & 0.4410 & 0.5033 \\
LPIPS  & 0.2750 & 0.2862 & 0.4165 \\
LC     & 0.3145 & 0.4151 & 0.4881 \\
DISC    & \textbf{0.4776}  & \textbf{0.4512} & \textbf{0.6203} \\
\bottomrule
\end{tabular}
\end{table}

\begin{table}[ht]
\centering
\small
\caption{OOD Type Classification Accuracy (\,$\mu \pm \sigma$\,) for \textbf{KMeans} (unsupervised Setting) and am \textbf{MLP} (Supervised Setting) by \emph{operator} and \emph{divergence/metric}. Best overall result in \textbf{bold}.}
\label{tab:traj_perf_pm}
\begin{tabular}{@{}llcc@{}}
\toprule
\textbf{Operator} & \textbf{Divergence / Metric} & \textbf{KMeans} & \textbf{MLP} \\
\midrule
\multicolumn{4}{l}{\emph{Aggregate}}\\
\quad DISC & --- & $\mathbf{0.4776 \pm 0.0005}$ & $\mathbf{0.7209 \pm 0.0173}$ \\
\midrule
\multicolumn{4}{l}{\emph{Intensity statistics}}\\
\quad Histogram & KL            & $0.3754 \pm 0.0023$ & $0.3769 \pm 0.0229$ \\
\quad Histogram & JS            & $0.4339 \pm 0.0015$ & $0.4582 \pm 0.0144$ \\
\quad Histogram & Total variation & $0.4288 \pm 0.0001$ & $0.4579 \pm 0.0108$ \\
\quad Histogram & Wasserstein   & $0.4286 \pm 0.0003$ & $0.4248 \pm 0.0150$ \\
\quad Vector    & Euclidean     & $0.4060 \pm 0.0007$ & $0.4061 \pm 0.0303$ \\
\quad Vector    & Cosine        & $0.3725 \pm 0.0022$ & $0.3866 \pm 0.0197$ \\
\midrule
\multicolumn{4}{l}{\emph{LBP (texture)}}\\
\quad LBP       & KL            & $0.3754 \pm 0.0023$ & $0.3769 \pm 0.0229$ \\
\quad LBP       & JS            & $0.4339 \pm 0.0015$ & $0.4582 \pm 0.0144$ \\
\quad LBP       & Total variation & $0.4288 \pm 0.0001$ & $0.4579 \pm 0.0108$ \\
\quad LBP       & Wasserstein   & $0.4286 \pm 0.0003$ & $0.4248 \pm 0.0150$ \\
\midrule
\multicolumn{4}{l}{\emph{HOG (shape/orientation)}}\\
\quad HOG       & KL            & $0.2673 \pm 0.0000$ & $0.2747 \pm 0.0177$ \\
\quad HOG       & JS            & $0.2653 \pm 0.0012$ & $0.2749 \pm 0.0187$ \\
\quad HOG       & Total variation & $0.2679 \pm 0.0004$ & $0.2744 \pm 0.0170$ \\
\quad HOG       & Wasserstein   & $0.2650 \pm 0.0002$ & $0.2786 \pm 0.0157$ \\
\midrule
\multicolumn{4}{l}{\emph{DTCWT (multi-scale wavelets)}}\\
\quad Level 1   & KL            & $0.3572 \pm 0.0018$ & $0.4777 \pm 0.0203$ \\
\quad Level 1   & MAD           & $0.2749 \pm 0.0007$ & $0.2891 \pm 0.0166$ \\
\quad Level 2   & KL            & $0.3342 \pm 0.0004$ & $0.3443 \pm 0.0184$ \\
\quad Level 2   & MAD           & $0.2725 \pm 0.0005$ & $0.2766 \pm 0.0183$ \\
\quad Level 3   & KL            & $0.2612 \pm 0.0015$ & $0.2593 \pm 0.0200$ \\
\quad Level 3   & MAD           & $0.2727 \pm 0.0011$ & $0.2680 \pm 0.0230$ \\
\quad LL band   & KL            & $0.2775 \pm 0.0007$ & $0.2621 \pm 0.0176$ \\
\quad LL band   & MAD           & $0.3580 \pm 0.0015$ & $0.3610 \pm 0.0143$ \\
\bottomrule
\end{tabular}
\end{table}

\paragraph{Why we adopt KL divergence.}
We compare several discrepancy choices:\emph{KL}, \emph{JS}, \emph{total variation}, \emph{Wasserstein}, and an absolute-deviation proxy (MAD). Table \ref{tab:traj_perf_pm} makes two points clear. \emph{First}, the best divergence is operator-dependent: LBP favors JS/TVD, while DTCWT (especially Level~1) prefers KL, and vector distances behave best with Euclidean. \emph{Second}, aggregating complementary operators (“DISC”) achieves best performance in unsupervised and supervised settings. We therefore standardize on \textbf{KL} in the main pipeline for distributional comparisons: it yields stable, monotone behaviour along the diffusion trajectory and competitive accuracy across families, while still allowing operator-specific alternatives (JS/TVD/Wasserstein/MAD) when robustness or symmetry is desired.

\paragraph{Ablation on the Number of Timesteps.} To study the contribution of different diffusion noise levels, we conducted an ablation on ImageNet that compares the full diffusion trajectory against restricted temporal subsets (early versus late timesteps) as well as the best-performing single timestep. Here, ``Early Half'' denotes features extracted from the first half of the diffusion trajectory, whereas ``Late Half'' denotes features from the second half.

As shown in Table~\ref{tab:timestep_ablation}, using the full trajectory consistently yields the best performance in both clustering and supervised evaluation. This supports our hypothesis that the temporal evolution of the diffusion process carries important diagnostic information that is not fully captured by any restricted subset of timesteps. Moreover, relying on a single timestep---even the best-performing one at $t=2500$---results in a marked performance drop, ranging from 3.8 to 8.5 percentage points. These results underscore the importance of aggregating information across the full diffusion trajectory.


\begin{table}[h!]
\centering
\caption{Ablation of the diffusion timesteps used for feature extraction on ImageNet. Using the full trajectory gives the strongest performance in both clustering and supervised evaluation.}
\label{tab:timestep_ablation}
\begin{tabular}{lcc}
\toprule
\textbf{Scenario} & \textbf{Clust. Acc} & \textbf{Sup. Acc} \\
\midrule
Full Trajectory & 49.1\% & 60.6\% \\
Early Half      & 47.8\% & 58.0\% \\
Late Half       & 47.0\% & 59.9\% \\
Best Single Step ($t=2500$) & 47.5\% & 56.8\% \\
\bottomrule
\end{tabular}
\end{table}

\vspace{0.5em}
\subsection{Tabular Data}
\label{sec:add_results:tabular}

\paragraph{Average performance and rank.}
Table~\ref{tab:transposed_results} summarizes mean AUROC and mean rank across tabular datasets. \textbf{DISC} attains the best average performance (\(0.860\)) and the best (lowest) average rank (\(4.54\)). Diffusion-enhanced baselines follow, with classical detectors trailing. The gap is not uniform—some classical methods excel on specific domains (e.g., network intrusion, small-\(d\) tabular).

\paragraph{Per-dataset results.}
Table~\ref{tab:baseline_performance} reports per-dataset AUROCs. DISC is consistently competitive and often state-of-the-art. On \texttt{http}/\texttt{WBC}, classical one-class methods approach saturation, narrowing the margin. We observe weaker gains on \texttt{Wilt} and \texttt{landsat}, where low SNR and class imbalance limit the added value of trajectory features; these are promising targets for feature-engineering or per-dataset calibration.

\begin{figure*}[htbp]
  \centering
  \begin{subfigure}[t]{0.32\textwidth}\centering
    \includegraphics[width=\linewidth]{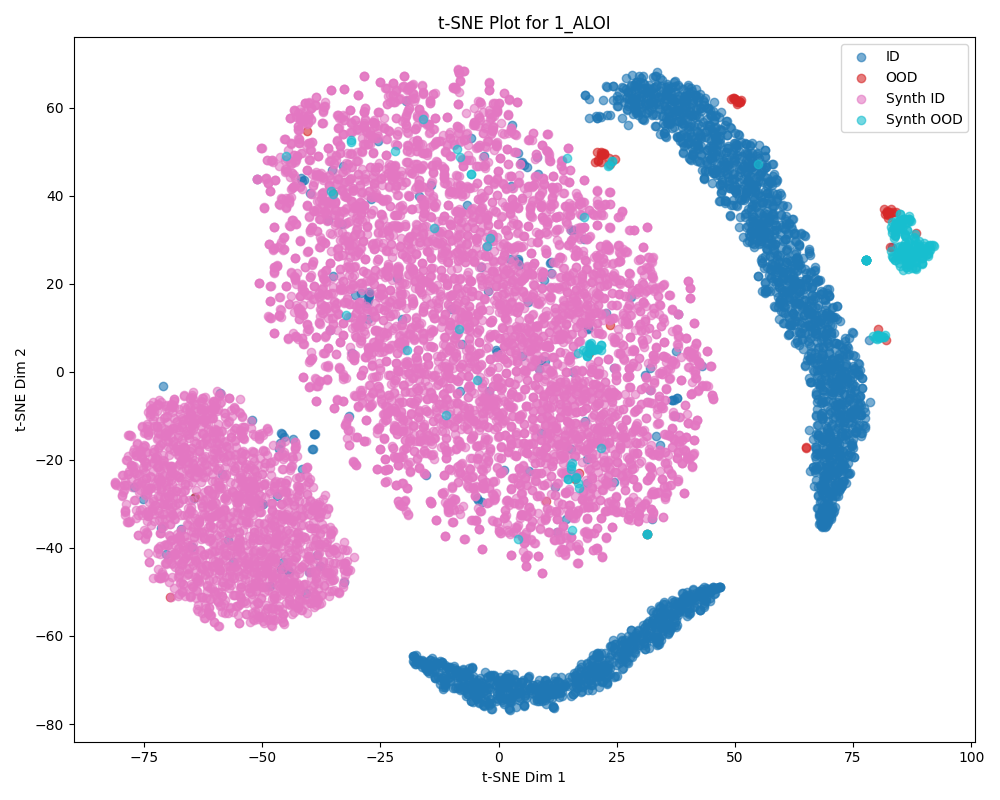}\caption{ALOI}\end{subfigure}\hfill
  \begin{subfigure}[t]{0.32\textwidth}\centering
    \includegraphics[width=\linewidth]{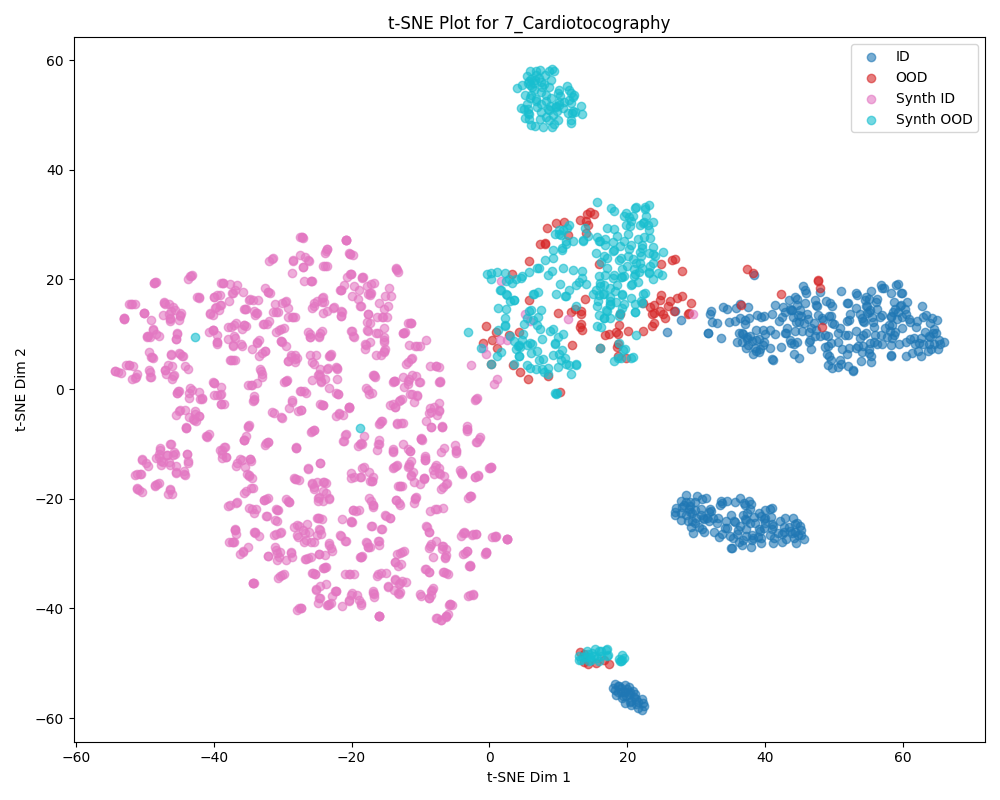}\caption{Cardiotocography}\end{subfigure}\hfill
  \begin{subfigure}[t]{0.32\textwidth}\centering
    \includegraphics[width=\linewidth]{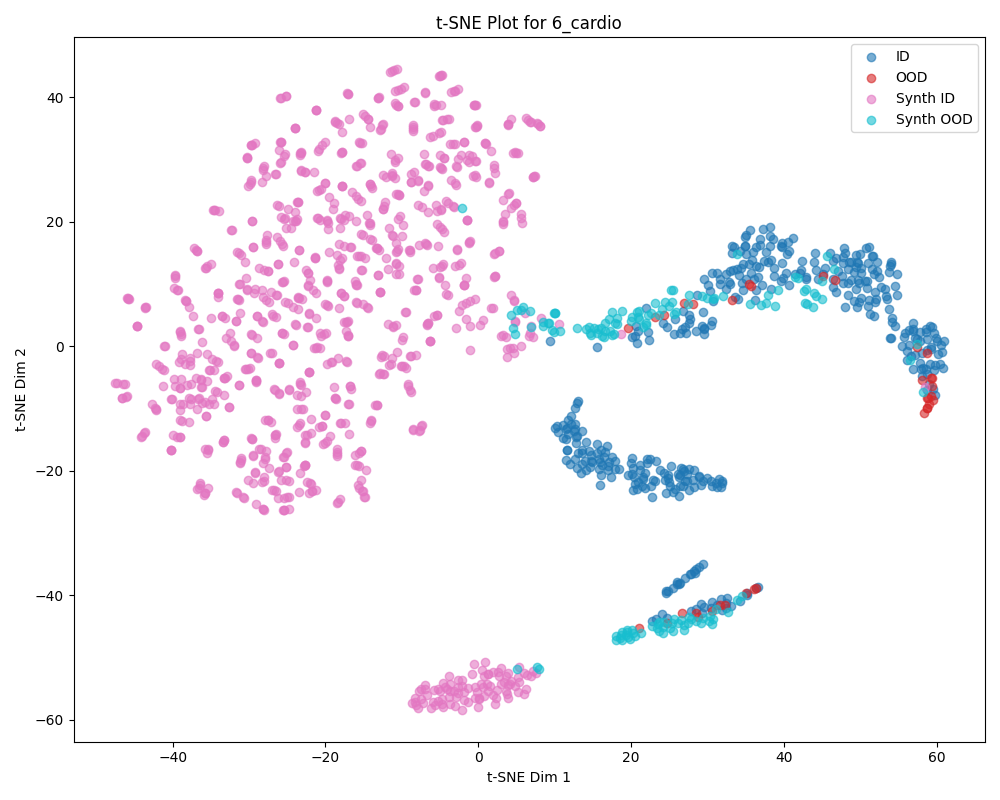}\caption{Cardio}\end{subfigure}

    \begin{subfigure}[t]{0.32\textwidth}\centering
    \includegraphics[width=\linewidth]{AISTATS2026PaperPack/Figures/tab_data/tsne_figs/RealSynthSplits_1_ALOI.png}\caption{ALOI}\end{subfigure}\hfill
  \begin{subfigure}[t]{0.32\textwidth}\centering
    \includegraphics[width=\linewidth]{AISTATS2026PaperPack/Figures/tab_data/tsne_figs/RealSynthSplits_7_Cardiotocography.png}\caption{Cardiotocography}\end{subfigure}\hfill
  \begin{subfigure}[t]{0.32\textwidth}\centering
    \includegraphics[width=\linewidth]{AISTATS2026PaperPack/Figures/tab_data/tsne_figs/RealSynthSplits_6_cardio.png}\caption{Cardio}\end{subfigure}

    \begin{subfigure}[t]{0.32\textwidth}\centering
    \includegraphics[width=\linewidth]{AISTATS2026PaperPack/Figures/tab_data/tsne_figs/RealSynthSplits_1_ALOI.png}\caption{ALOI}\end{subfigure}\hfill
  \begin{subfigure}[t]{0.32\textwidth}\centering
    \includegraphics[width=\linewidth]{AISTATS2026PaperPack/Figures/tab_data/tsne_figs/RealSynthSplits_7_Cardiotocography.png}\caption{Cardiotocography}\end{subfigure}\hfill
  \begin{subfigure}[t]{0.32\textwidth}\centering
    \includegraphics[width=\linewidth]{AISTATS2026PaperPack/Figures/tab_data/tsne_figs/RealSynthSplits_6_cardio.png}\caption{Cardio}\end{subfigure}
  \caption{t-SNE of DISC features on tabular datasets. Blue: ID, red: real OOD, magenta: synthetic ID, cyan: synthetic OOD. Synthetic perturbations separate cleanly from ID; real OODs are closer yet remain distinguishable.}
  \label{fig:tabular_tsne}
\end{figure*}

\paragraph{t-SNE examples and real vs.\ synthetic disentanglement.}
Figure~\ref{fig:tabular_tsne} visualizes DISC embeddings. Across datasets, synthetic OODs form compact, well-separated clusters (cyan), while real OODs (red) sit closer to, but still distinct from, the ID manifold (blue). The improved separation over synthetic perturbations indicates that diffusion trajectories capture shifts aligned with density-preserving transformations, while still retaining sensitivity to semantic deviations in real OODs. (As before, t-SNE distances are not calibrated across panels.)

\begin{table}[ht]
\centering
\resizebox{\textwidth}{!}{
\begin{tabular}{l*{17}{c}}
\toprule
Metric & DISC (ours) & DDPM(MSE-IFOREST) & CBLOF & IForest & HBOS & PCA & KNN & OCSVM & SOD & ECOD & COPOD & DDPM(MSE) & COF & DDPM(MSE-GEV) & LOF & DDPM(MSE-GPD) & DAGMM \\
\midrule
avg\_performance
  & 0.860246 & 0.849808 & 0.765023 & 0.763110 & 0.737901 & 0.729835 & 0.708433 & 0.695064 & 0.693449 & 0.667007 & 0.665289 & 0.641453 & 0.627385 & 0.621617 & 0.620708 & 0.608327 & 0.522257 \\
avg\_rank
  & 4.538462 & 5.076923 & 6.615385 & 6.615385 & 8.000000 & 8.025641 & 7.692308 & 9.461538 & 9.461538 & 8.858974 & 9.166667 & 11.371795 & 11.410256 & 11.730769 & 10.794872 & 11.769231 & 12.410256 \\
\bottomrule
\end{tabular}
}

\caption{Average AUROC and average rank over tabular datasets. Lower rank is better.}
\label{tab:transposed_results}
\end{table}

\begin{table}[ht]
  \centering
  \small
  \resizebox{\textwidth}{!}{

  \begin{tabular}{lrrrrrrrrrrrrr}
    \toprule
     & DISC  & IForest & OCSVM & CBLOF & COF & COPOD & ECOD & HBOS & KNN & LOF & PCA & SOD & DAGMM \\
    \midrule
    1\_ALOI              & 0.965 & 0.567 & 0.558 & 0.552 & 0.647 & 0.537 & 0.563 & 0.526 & 0.615 & 0.666 & 0.566 & 0.611 & 0.520 \\
    4\_breastw           & 0.919 & 0.983 & 0.803 & 0.968 & 0.388 & 0.997 & 0.992 & 0.989 & 0.970 & 0.406 & 0.951 & 0.940 & NaN   \\
    5\_campaign          & 0.991 & 0.717 & 0.657 & 0.642 & 0.574 & 0.783 & 0.767 & 0.785 & 0.723 & 0.590 & 0.728 & 0.692 & 0.560 \\
    6\_cardio            & 0.712 & 0.932 & 0.939 & 0.899 & 0.714 & 0.923 & 0.936 & 0.847 & 0.766 & 0.663 & 0.956 & 0.732 & 0.750 \\
    7\_Cardiotog         & 0.971 & 0.676 & 0.779 & 0.645 & 0.538 & 0.670 & 0.784 & 0.609 & 0.562 & 0.595 & 0.747 & 0.517 & 0.620 \\
    8\_celeba            & 1.000 & 0.704 & 0.707 & 0.740 & 0.386 & 0.757 & 0.765 & 0.762 & 0.596 & 0.386 & 0.794 & 0.479 & 0.447 \\
    9\_census            & 0.749 & 0.595 & 0.549 & 0.603 & 0.416 & NaN   & NaN   & 0.649 & 0.669 & 0.474 & 0.687 & 0.621 & 0.597 \\
    10\_cover            & 0.955 & 0.867 & 0.926 & 0.893 & 0.769 & 0.886 & 0.919 & 0.802 & 0.860 & 0.846 & 0.937 & 0.745 & 0.899 \\
    11\_donors           & 0.999 & 0.777 & 0.724 & 0.620 & 0.708 & 0.818 & 0.890 & 0.784 & 0.819 & 0.589 & 0.831 & 0.560 & 0.708 \\
    12\_fault            & 0.980 & 0.570 & 0.477 & 0.641 & 0.621 & 0.439 & 0.454 & 0.513 & 0.730 & 0.589 & 0.460 & 0.681 & 0.459 \\
    14\_glass            & 1.000 & 0.771 & 0.354 & 0.829 & 0.722 & 0.724 & 0.658 & 0.772 & 0.823 & 0.692 & 0.663 & 0.734 & 0.761 \\
    16\_http             & 0.965 & 1.000 & 0.996 & 0.996 & 0.888 & 0.993 & 0.981 & 0.995 & 0.034 & 0.275 & 0.997 & 0.780 & NaN   \\
    17\_InternetAds      & 0.906 & 0.690 & 0.683 & 0.706 & 0.638 & 0.671 & 0.671 & 0.680 & 0.700 & 0.658 & 0.617 & 0.619 & NaN   \\
    18\_Ionosphere       & 0.884 & 0.845 & 0.759 & 0.907 & 0.868 & 0.793 & 0.732 & 0.625 & 0.883 & 0.906 & 0.792 & 0.864 & 0.734 \\
    19\_landsat          & 0.738 & 0.476 & 0.361 & 0.635 & 0.535 & 0.415 & 0.361 & 0.551 & 0.580 & 0.539 & 0.358 & 0.595 & 0.439 \\
    41\_Waveform         & 0.930 & 0.715 & 0.563 & 0.724 & 0.726 & 0.750 & 0.624 & 0.688 & 0.738 & 0.733 & 0.655 & 0.686 & 0.493 \\
    42\_WBC              & 0.808 & 0.990 & 0.990 & 0.995 & 0.609 & 0.991 & 0.991 & 0.987 & 0.906 & 0.542 & 0.982 & 0.946 & NaN   \\
    44\_Wilt             & 0.659 & 0.419 & 0.313 & 0.325 & 0.497 & 0.334 & 0.363 & 0.325 & 0.484 & 0.507 & 0.204 & 0.533 & 0.373 \\
    47\_yeast            & 0.997 & 0.378 & 0.410 & 0.448 & 0.445 & 0.370 & 0.438 & 0.396 & 0.391 & 0.453 & 0.412 & 0.425 & 0.411 \\
    \bottomrule
  \end{tabular}
  }
  \caption{Per-dataset AUROC for DISC and baselines (DDPM variants removed for compactness).}
  \label{tab:baseline_performance}
\end{table}

\vfill


\end{document}